\documentclass{article}
\usepackage{ijcai24}

\usepackage{times}
\usepackage{soul}
\usepackage{url}
\usepackage[hidelinks]{hyperref}
\usepackage[utf8]{inputenc}
\usepackage[small]{caption}
\usepackage{graphicx}
\usepackage{amsmath}
\usepackage{amsthm}
\usepackage{booktabs}
\usepackage{algorithm}
\usepackage{algorithmic}
\usepackage[switch]{lineno}

\usepackage{colortbl}  
\usepackage{xcolor}
\usepackage{array}   
\usepackage{multirow}
\usepackage{subcaption}
\usepackage{amssymb}
\usepackage{colortbl}  

\newtheorem{theorem}{Theorem}

\title{Hacking Task Confounder in Meta-Learning}

\author{
Jingyao Wang$^{1,2}$ 
\and
Yi Ren$^{1}$
\and
Zeen Song$^{1,2}$
\and
Jianqi Zhang$^{1,2}$
\and\\
Changwen Zheng$^{1}$
\And
Wenwen Qiang$^{1,2}$\footnote{Corresponding Author}\\
\affiliations
$^1$Institute of Software Chinese Academy of Sciences\\
$^2$University of Chinese Academy of Sciences\\
\emails
\{wangjingyao23, renyi, songzeen, zhangjianqi, changwen, qiangwenwen\}@iscas.ac.cn
}

\begin{document}

\maketitle

\begin{abstract}
    Meta-learning enables rapid generalization to new tasks by learning knowledge from various tasks. It is intuitively assumed that as the training progresses, a model will acquire richer knowledge, leading to better generalization performance. However, our experiments reveal an unexpected result: there is negative knowledge transfer between tasks, affecting generalization performance. To explain this phenomenon, we conduct Structural Causal Models (SCMs) for causal analysis. Our investigation uncovers the presence of spurious correlations between task-specific causal factors and labels in meta-learning. Furthermore, the confounding factors differ across different batches. We refer to these confounding factors as ``Task Confounders". Based on these findings, we propose a plug-and-play Meta-learning Causal Representation Learner (MetaCRL) to eliminate task confounders. It encodes decoupled generating factors from multiple tasks and utilizes an invariant-based bi-level optimization mechanism to ensure their causality for meta-learning. Extensive experiments on various benchmark datasets demonstrate that our work achieves state-of-the-art (SOTA) performance. The code is provided in \url{https://github.com/WangJingyao07/MetaCRL}.
\end{abstract}



\section{Introduction}
\label{sec:1}

Meta-learning aims to develop models that can be rapidly transferred to previously unseen tasks. To achieve this, it first learns from diverse tasks to obtain models with high learning capacities. Then, it fine-tunes these models with little data from unseen tasks to obtain the desired ones. Recently, meta-learning has been widely applied in various fields, e.g., affective computing ~\cite{li2023compound}, image classification ~\cite{qiang2023meta}, and robotics ~\cite{schrum2022mind}.

During the training phase, each batch consists of a series of randomly sampled $N$-way $K$-shot tasks, where $N$ denotes the number of classes per task and $K$ denotes the number of samples per class. The samples in each task are divided into a support set and a query set. Then, meta-learning models are trained in a bi-level optimization manner ~\cite{wang2021bridging,wang2023awesome}. In brief, at the first level, the desired model for each task is fine-tuned by training on the support set using the meta-learning model. At the second level, the meta-learning model is learned using the query sets from all training tasks and the corresponding expected models for each task.
Therefore, a widely adopted hypothesis is that as training progresses, the meta-learning model will acquire richer knowledge that can be transferred well to downstream tasks, achieving better performance ~\cite{rivolli2022meta}. 

\begin{figure}
    \centering
    \includegraphics[width=\linewidth]{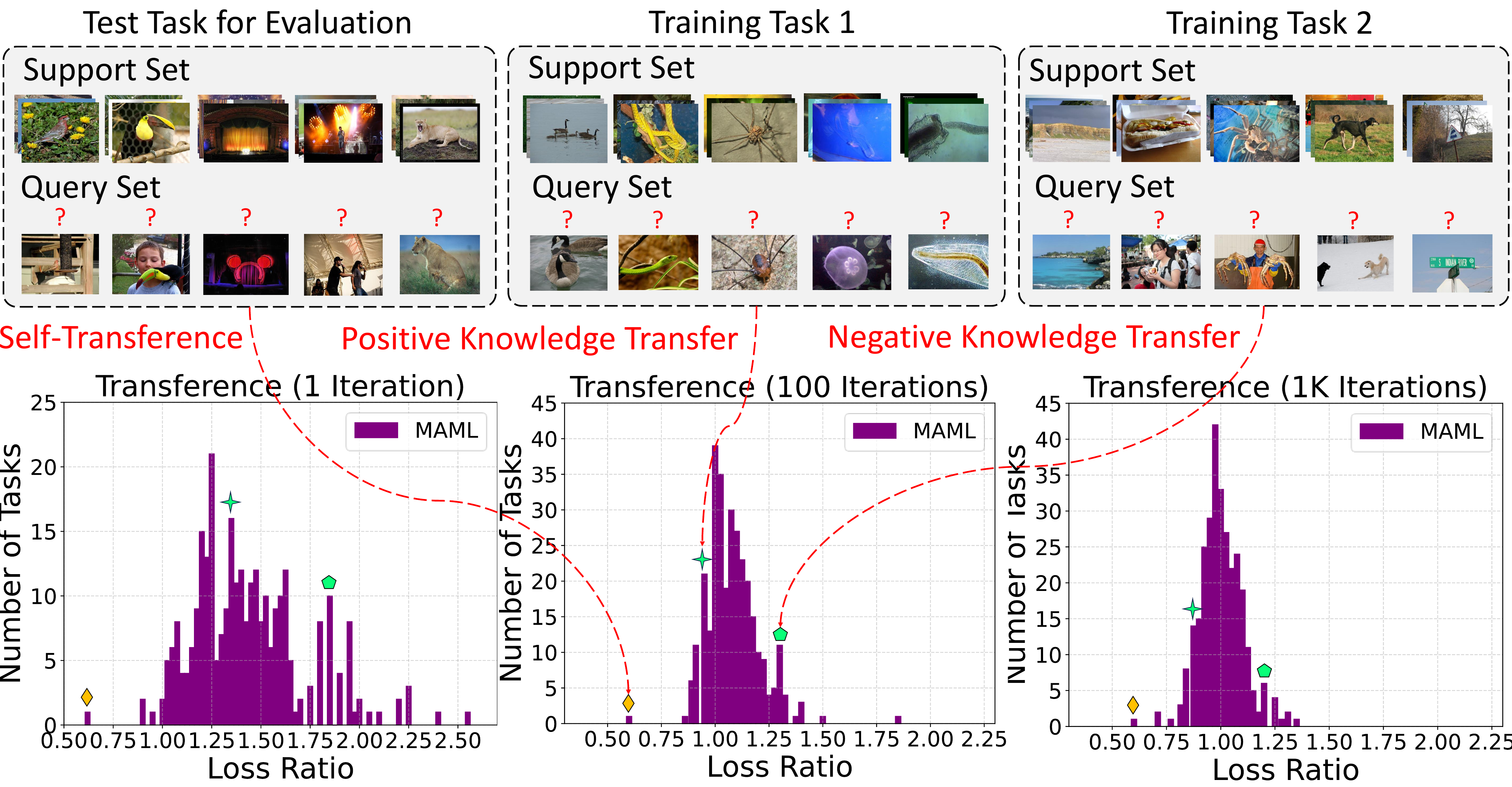}
    \caption{Knowledge transfer to a specific test task. For both positive knowledge transfer ($\mathcal{R}_{i,j}<1$) and negative knowledge transfer ($\mathcal{R}_{i,j}>1$), an exemplar task is shown. Here, we simply use the $\mathcal{R}_{i,j}$ threshold to classify the knowledge transfer as positive or negative. See Subsection \ref{sec:3.2} and Appendix F for more details.}
    \label{fig:intro}
\end{figure}

\begin{figure}
    \centering
    \begin{subfigure}{0.49\columnwidth}
        \centering
        \includegraphics[width=0.85\columnwidth]{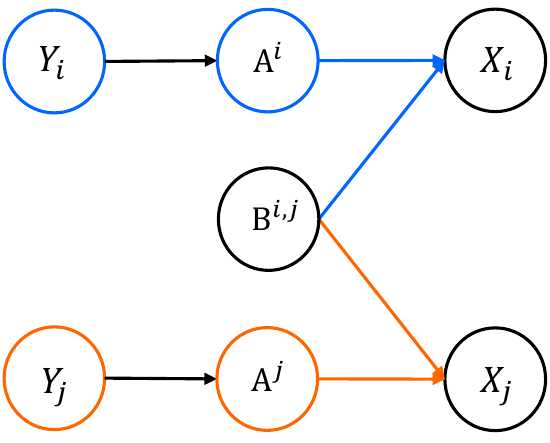}
        \caption{data generation mechanism} 
        \label{fig2.1}
    \end{subfigure}
    \hfill
    \begin{subfigure}{0.49\columnwidth}
        \centering
        \includegraphics[width=0.85\columnwidth]{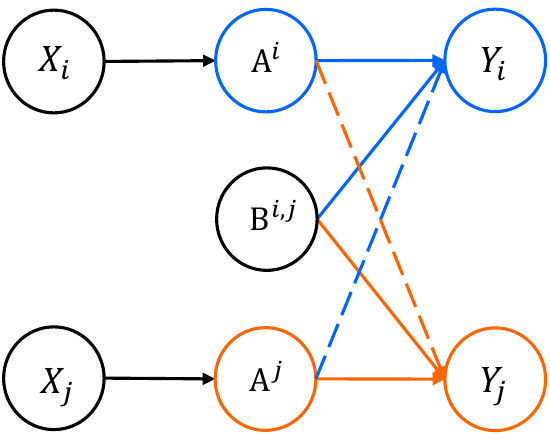}
        \caption{meta-learning process} 
        \label{fig2.2}
    \end{subfigure}
    \caption{Structural Causal Models (SCM) regarding two tasks $\tau_i$ and $\tau_j$, where $(X_i, Y_i)$ and $(X_j, Y_j)$ are the samples and corresponding labels of these tasks. The solid line means the true causal correlation, and the dotted line means the spurious correlation. (a) is constructed based on the ground-truth causal mechanism, while (b) can be viewed as the inverse process of the generating mechanism.}
    \label{fig2}
\end{figure}

However, our toy experiments reveal a conflicting phenomenon, i.e., the knowledge learned from the training tasks may be harmful to the unseen test tasks (See Subsection \ref{sec:3.2} for more details). 
Specifically, we first randomly sample 400 tasks from miniImageNet dataset~\cite{vinyals2016matching} and divide them into a training set and a test set.
Then, we define a metric $\mathcal{R}_{i,j}$ to evaluate whether the meta-learning model trained on the training tasks can perform better on the test task, i.e., quantify the knowledge transfer performance from the training tasks to each test task. If $\mathcal{R}_{i,j}<1$, the learned knowledge from the training task can help improve the model performance on the test task (positive knowledge transfer), while $\mathcal{R}_{i,j}>1$ implies the learned knowledge is harmful to the test task (negative knowledge transfer). We use MAML~\cite{finn2017model} as the baseline and record the score of $\mathcal{R}_{i,j}$ in the middle of training ~\cite{fifty2020measuring,multimodalmeta}. Figure \ref{fig:intro} shows the results. Ideally, all the knowledge transfer between tasks should be positive, i.e., $\mathcal{R}_{i,j}<1$. The results show that there always exists negative knowledge transfer between tasks.

To explore the reasons behind this phenomenon, we propose using causal theory for analysis (See Subsection \ref{sec:3.3} for details). We begin by constructing Structural Causal Models (SCMs) for the training phase of ML, as shown in Figure \ref{fig2}. 
In the SCMs, ${\rm{A}}^{i}$ and ${\rm{A}}^{j}$ are the distinct causal factors of task $\tau_i$ and task $\tau_j$, and ${\rm{B}}^{i,j}$ means the shared causal factors of these two tasks. Meanwhile, causal factors can be considered as different semantics of the data, e.g., color and shape, also considered as generating factors used for data generation \cite{zimmermann2021contrastive}. 
Since meta-learning performs joint learning on all the training tasks, it acquires all the causal factors.
Thus, the non-overlapping causal factors ${\rm{A}}^{i}$ of $\tau_i$ may cause spurious correlations with $\tau_j$, and ${\rm{A}}^{j}$ holds the same with $\tau_i$. These misleading correlations between training tasks will introduce bias into the learned knowledge and ultimately affect generalization, which is called ``\textbf{task confounder}''.

To address this issue, we propose a plug-and-play meta-learning causal representation learner (MetaCRL) to encode decoupled causal knowledge, thereby eliminating task confounders. It consists of two modules: the disentangling module and the causal module. The former aims to extract generating factors across all tasks and provide a subset of factors relevant to each task, while the latter is responsible for ensuring their causality. The modules achieve their objectives through a simple bi-level optimization mechanism with regularization terms. By incorporating MetaCRL into meta-learning, we dynamically eliminate task confounders during the meta-training process. Through extensive evaluations of multiple meta-learning benchmarks, we demonstrate that MetaCRL can significantly improve performance.

In summary, our contributions are as follows: 
\begin{itemize}
    \item We discover a counterintuitive phenomenon: there is negative knowledge transfer between tasks, resulting in reduced model generalization performance.
    \item We construct an SCM to analyze the phenomenon with causal theory, finding spurious correlations, named ``Task Confounders'', between non-shared causal factors of the meta-training tasks and the label space.
    \item We propose MetaCRL, a plug-and-play meta-learning causal representation learner to eliminate task confounders, thus improving generalization performance.
    \item Extensive experiments on various scenarios demonstrate the outstanding performance of our MetaCRL.
\end{itemize}



\section{Related Work}
\label{sec:2}

\paragraph{Meta-learning} aims to learn general knowledge from various training tasks, and then generalize to new tasks based on the acquired knowledge. Typical methods can be categorized into two types: optimization-based ~\cite{finn2017model,nichol2018reptile,guo2024self} and metric-based ~\cite{snell2017prototypical,sung2018learning,chen2020variational} methods. They both rely on shared structures and bi-level learning mechanisms to learn general knowledge, resulting in remarkable performance on new tasks. However, meta-learning still faces the crisis of performance degradation. Various approaches have been proposed to address this issue, such as adding adaptive noise~\cite{lee2020meta}, reducing inter-task disparities~\cite{jamal2019task}, limiting the trainable parameters~\cite{yin2019meta,oh2020boil}, and task augmentation~\cite{yao2021improving}. Despite alleviating performance degradation, they ignore the interaction between tasks, which is shown to be crucial in Section \ref{sec:3}. In this study, we analyze the knowledge transfer effects between different training tasks with causal theory, and focus on the fundamental causes of performance degradation in meta-learning.

\paragraph{Causal learning} aims to explore the causal relationships between variables in machine learning, modeling the target with a directed acyclic graph, also known as a causal model.
It has been shown to aid models in unearthing underlying causal factors ~\cite{yang2021deconfounded,zhang2020causal,nogueira2022methods}. Current research attempts to combine causal knowledge with meta-learning methods to address domain challenges. Yue et al. ~\cite{yue2020interventional} removed performance limitations of pre-trained knowledge through backdoor regulation. Ton et al. ~\cite{ton2021meta} utilized causal knowledge to distinguish causes and effects in a bivariate environment with limited data. Jiang et al. ~\cite{jiang2022role} used causal graphs to remove undesirable memory effects. While they all combine meta-learning and causal learning, their focus is on addressing problems that differ from ours.


\section{Problem Formulation and Analysis}
\label{sec:3}
In this section, we first present the notation and problem definition of meta-learning. Next, we conduct experiments to evaluate the interaction between different tasks and illustrate the empirical evidence, i.e., the knowledge learned from the training tasks may be harmful to the unseen test tasks, reducing generalization performance. Finally, we construct SCMs to explore the reasons behind the empirical evidence.

\subsection{Preliminaries}
\label{sec:3.1}
Given a task distribution $p(\mathcal{T})$, the meta-training dataset $\mathcal{D}_{tr}$ and the meta-test dataset $\mathcal{D}_{te}$ are all sampled from $p(\mathcal{T})$ without class-level overlap. During the training phase of ML, each batch contains $N_{tr}$ tasks, denoted as $\left \{ \tau_i \right \}_{i=1}^{N_{tr}} \in \mathcal{D}_{tr}$, and each task $\tau_{i}$ consists of a support set $\mathcal{D}_i^s=(X_i^s,Y_i^s)=\{ (x_{i,j}^s,y_{i,j}^s)  \}_{j=1}^{N_i^s}$ and a query set $\mathcal{D}_i^q=(X_i^q,Y_i^q)=  \{ (x_{i,j}^q,y_{i,j}^q)  \}_{j=1}^{N_i^q}$, where $(x_{i,j}^{\cdot},y_{i,j}^{\cdot})$ represents the sample and the corresponding label, and $N_i^{\cdot}$ denotes the number of the samples. The meta-learning model $f_{\theta}=h \circ g$ utilizes the feature encoder $g$ and the classifier $h$ to learn the above tasks. 

The learning mechanism of meta-learning is regarded as a bi-level optimization process. At the first level, it fine-tune the desired model $f_{\theta}^i$ for task $\tau_i$ by training on the support set $\mathcal{D}_i^s$ using the meta-learning model $f_{\theta}$, presented as:
\begin{equation}
\label{eq:inner}
\begin{array}{l}
    \qquad \qquad f^i_{\theta} \gets f_{\theta} -\alpha \nabla_{f_{\theta}}\mathcal{L}(Y_i^s,X_i^s,f_{\theta}) \\[5pt]
    s.t. \quad \mathcal{L}(Y_i^s,X_i^s,f_{\theta} )=\frac{1}{N_i^s} \sum_{j=1}^{N_i^s}y_{i,j}^s\log {f_\theta }(x_{i,j}^s)
\end{array}
\end{equation}
where $\alpha$ is the learning rate. At the second level, the meta-learning model $f_{\theta}$ is learned using the query sets $\mathcal{D}^q$ from all training tasks and the expected models for each task:
\begin{equation}
\label{eq:outer}
\begin{array}{l}
    \qquad f_{\theta} \gets f_{\theta}-\beta \nabla_{f_{\theta}}\frac{1}{N_{tr}}\sum_{i=1}^{N_{tr}}\mathcal{L}(Y_i^q,X_i^q,f_\theta ^i)  \\[5pt]
    s.t. \quad \mathcal{L}(Y_i^q,X_i^q,f_\theta ^i)= \frac{1}{N_i^q} \sum_{j=1}^{N_i^q}y_{i,j}^q\log {f_\theta^i }(x_{i,j}^q) 
\end{array}
\end{equation}
where $\beta$ is the learning rate. Note that $f_\theta^i$ is obtained by taking the derivative of $f_\theta$, so $f_\theta^i$ can be regarded as a function of $f_\theta$. Therefore, the update of $f_\theta$ mentioned in Eq.\ref{eq:outer} can be viewed as calculating the second derivative of $f_\theta$.

\subsection{Empirical Evidence}
\label{sec:3.2}

From above and \cite{wang2021bridging}, meta-training on one batch can be viewed as a multi-task learning process. Meanwhile, a well-learned model should contain knowledge of all training tasks. Therefore, intuitively, one might assume that as training progresses, the meta-learning model will acquire richer knowledge (related to all tasks) and transfer better to downstream tasks, achieving great generalization. However, our toy experiments reveal that this is not always true.

Before introducing the toy experiments, we first present a method to quantify the influence of transferring knowledge learned from one task to the target task. 
For task $\tau_i$, the model $f_{\theta}$ uses the support set $\mathcal{D}_i^s$ to obtain $f_{\theta}^i$ via Eq.\ref{eq:inner}. Here, $f_{\theta}^i$ is considered to integrate the knowledge of task $\tau_i$ into $f_{\theta}$. Then, for task $\tau_j$, we first obtain the model $f_{\theta}^{j,1}$ by training $f_{\theta}^i$ on the support set $\mathcal{D}_j^s$, and then obtain the model $f_{\theta}^{j,2}$ by training $f_{\theta}$ on $\mathcal{D}_j^s$.
Next, we calculate their losses on the query set $\mathcal{D}_j^q$, expressed as $\mathcal{L}(\mathcal{D}_j^q,f_{\theta}^{j,1}) $ and $\mathcal{L}(\mathcal{D}_j^q,f_{\theta}^{j,2}) $, respectively.
Finally, we calculate the ratio between these two losses, denoted as $\mathcal{R}_{i,j}$, which quantifies the performance of knowledge transfer from task $\tau_i$ to task $\tau_j$. Thus, we have:
\begin{equation}
\label{eq:kr}
\begin{array}{l}
    \mathcal{R}_{i,j}=\frac{\mathcal{L}(\mathcal{D}_j^q,f_{\theta}^{j,1})}{\mathcal{L}(\mathcal{D}_j^q,f_{\theta}^{j,2}) } 
\end{array}
\end{equation}
if $\mathcal{R}_{i,j}<1$, it means that task $\tau_i$ has a positive knowledge transfer effect on task $\tau_j$. On the other hand, if $\mathcal{R}_{i,j}>1$, it indicates the negative knowledge transfer effect of $\tau_i$ on $\tau_j$.

Next, we conduct experiments based on the quantitative method described above. We first randomly sample 400 tasks from miniImageNet dataset, which are divided into a training set of 300 tasks and a test set of 100 tasks. Then, we use MAML as the baseline to calculate the score of $\mathcal{R}_{i,j}$ from the training tasks to each test task in the middle of training. 

Figure \ref{fig:intro} shows the histograms of the knowledge transfer in the training phase of meta-learning along with exemplar tasks. From the results, we observe that as training proceeds, although the knowledge transfer effects become more and more positive, there always exists negative knowledge transfer between different tasks. It indicates that the training process of meta-learning cannot always obtain effective knowledge for unseen test tasks, and the aforementioned intuitive hypothesis is limited. 
Note that we also conduct experiments under various different settings, including using multiple meta-learning baselines, using different datasets, and training on multiple tasks simultaneously (the effect of multiple training tasks to a single test task), the impact of negative knowledge transfer always exists. 
More details and the full results are provided in Appendix F.

\subsection{Causal Analysis and Motivation}
\label{sec:3.3}
To explore the reasons behind the above phenomenon, we propose using causal theory for analysis. We first construct a Structural Causal Model (SCM) based on the ground-truth causal mechanisms ~\cite{suter2019robustly,hu2022improving}, as shown in Figure \ref{fig2.1}. Specifically, this SCM contains two tasks $\tau_i$ and $\tau_j$, where $Y_i$ and $Y_j$ denote the label variables for tasks $\tau_i$ and $\tau_j$, $X_i$ and $X_j$ signify the corresponding generated samples for these two tasks, respectively. Meanwhile, ${{\rm A}^i}$ and ${{\rm A}^j}$ represent the distinct sets of causal factors specific to tasks $\tau_i$ and $\tau_j$, while ${{\rm B}^{i,j}}$ encompasses shared causal factors. 
In this SCM, we assume that the samples $X_i$ and $X_j$ are generated by disentangled causal mechanisms using the causal factors, then $p({X_i}|{{\rm A}^i},{{\rm{B}}^{i,j}}) = \prod\nolimits_k {p({X_i}|{{\rm A}^i_k})} \prod\nolimits_t {p({X_i}|{\rm B}_t^{i,j})}$, where ${\rm A}^i_k$ denotes the $k$-th factor of ${\rm A}^i$, and ${\rm B}_t^{i,j}$ denotes the $t$-th factor of ${\rm B}^{i,j}$. Since ${\rm A}^i$, ${\rm A}^j$, and ${\rm B}^{i,j}$ represent high-level knowledge of the data, we could naturally define the task label variable $Y_i$ for task $i$ as the cause of the ${\rm B}^{i,j}$ and ${\rm A}^i$. For the task $\tau_i$, we call ${\rm B}^{i,j}$ and ${\rm A}^i$ as the causal feature variables that are causally related to $Y_i$, and we call ${\rm A}^j$ as the non-causal feature variables to task $\tau_i$. Therefore, we have $p({X_i}|{{\rm A}^i},{{\rm{B}}^{i,j}},{{\rm A}^j}) = p({X_i}|{{\rm A}^i},{{\rm{B}}^{i,j}})$.

Based on the proposed SCM, an ideal meta-learning predictor for each task should only utilize causal factors and be invariant to any intervention on non-causal factors. However, the joint learning of multiple tasks in meta-learning could give rise to the issue of using non-causal factors for unseen tasks, also known as spurious correlations, thereby making it challenging to achieve optimal predictions. 
To verify this claim, 
we consider the scenario of two binary classification tasks for simple but clear explanations. Let $Y_i$ and $Y_j$ be variables from $\left\{ { \pm 1} \right\}$, we assume $\tau_i$ and $\tau_j$ have non-overlapping factors, i.e., ${{\rm{B}}^{i,j}} = \emptyset$, and the elements in ${{\rm A}^i}$ and ${{\rm A}^j}$ satisfy the constraint of Gaussian distribution. Then, we have:
\begin{theorem} \label{accepted}
If the correlation between $Y_i$ and $Y_j$ is not equal to 0.5, the optimal classifier has non-zero weights for non-causal factors for each task. If the correlation between $Y_i$ and $Y_j$ equals 0.5 with limited training data, the optimal classifier also has non-zero weights for non-causal factors in each task.
\end{theorem}

As inferred from the aforementioned theorem, the learned model leverages the causal factors from other tasks to facilitate the learning of the target task. Taking the task $\tau_i$ as an example, the meta-learning model uses the causal factors ${{\rm A}^j}$ belonging to the task $\tau_j$ for learning $Y_i$. Therefore, there is a spurious correlation between ${{\rm A}^j}$ and $Y_i$, which can be represented as a spurious path ${{\rm A}^j} \to Y_i$. Similarly, we can obtain the spurious path ${{\rm A}^i} \to Y_j$ for task $\tau_j$. These spurious correlations are called ``task confounders'', which are the reasons that lead to negative knowledge transfer in Subsection \ref{sec:3.2}. The learning process can be viewed as the inverse process of the generating mechanism. Therefore, we can obtain the SCM with two spurious paths as illustrated in Figure \ref{fig2.2}, which reflects the internal mechanism of task confounders in multi-task learning. The proof is provided in Appendix A.



\section{Methodology}
\label{sec:4}

Based on the above analysis, we know that task confounders cause spurious correlations between causal factors and labels. An ideal meta-learning model should identify knowledge that is causally related to each task and learn from the identified multi-task knowledge. 
Therefore, we propose MetaCRL, a plug-and-play meta-learning causal representation learner that can encode decoupled causal factors for more efficient ML. It consists of two modules: (i) the disentangling module which aims to extract generating factors and eliminate task confounders; and (ii) the causal module which aims to ensure the causality of the obtained generating factors.
In this section, we first introduce the disentangling module and the causal module in Subsections \ref{sec:4.1} and \ref{sec:4.2}, respectively. Next, we provide the overall objective in Subsection \ref{sec:4.3}. The pseudocode and pipeline of MetaCRL are shown in Appendix B.

\subsection{Disentangling Module}
\label{sec:4.1}
In this module, we aim to obtain the whole generating factors related to all tasks and the task-specific generating factors related to each single task. Specifically, we first obtain the whole generating factors by learning a semantic matrix $\Xi$. Next, we use a grouping function $f_{gr}$ to acquire subsets of generating factors relevant to every single task. Note that this module does not guarantee the causality of the obtained generating factors, which will be addressed in the causal module.

For a pre-trained encoder, different channels of the feature representations are related to different kinds of semantics~\cite{islam2020much}. Thus, we propose to use the feature representation to learn the generating factors. During the training phase, we denote the $N_{tr}$ training tasks as $\left \{ \tau_i \right \}_{i=1}^{N_{tr} }$. Suppose that the number of generating factors is $N_k$, then, we propose obtaining these $N_k$ factors through the learning of a matrix $\Xi \in {{\mathbb{R}}^{N_z \times N_k}}$. Here, $N_z$ represents the dimension of the feature representation, i.e., the output dimension of the encoder $g$, and each column of $\Xi$ represents a distinct factor. Based on $\Xi$, we can obtain a new representation of each sample, which can be called a generating representation, e.g., the generating representation for $x_{i,j}^s$ can be presented as ${\Xi ^{\rm{T}}}g(x_{i,j}^s)$.

Generally, generating factors in geometric space can be conceptualized as coordinate basis vectors, where each generating factor corresponds to a specific basis vector \cite{jensen2004semantics}. Moreover, different coordinate bases can undergo mutual transformations via a reversible matrix, implying their equivalence. Hence, learning a task-specific matrix, serving as a base matrix, allows us to approximate task-related generating factors. Therefore, for $\Xi$ to be considered a generating factor matrix, we need to constrain the column vectors of $\Xi$ to be orthogonal to each other. Then we have:
\begin{equation}
\label{eq:Ldm}
    \mathcal{L}_{\rm{DM}}(\Xi)=\sum\limits_{i = 1}^{{N_k} - 1}\sum\limits_{j = i + 1}^{N_k} {\Xi _{:,i}^{\rm{T}}{\Xi _{:,j}}} 
\end{equation}
where ${{\Xi _{:,i}}}$ represents the $i$-th column of $\Xi$. Minimizing $\mathcal{L}_{\rm{DM}}(\Xi)$ makes the different columns of $\Xi$ orthogonal to each other, thus leading $\Xi$ to be task-related generating factors.

Next, for all the $N_{tr}$ training tasks, the generating factors should be divided into $N_{tr}$ overlapping groups, and each group corresponds to a task. To obtain these groups, we propose a learnable grouping function $f_{gr}$, which is implemented using Multi-Layer Perceptrons (MLPs) to acquire task-specific generating factors. Take task $\tau_i$ as an example, we first calculate the average sample $x_i$ for this task, i.e., ${x_i} = \frac{1}{N_i^s+N_i^q} (\sum\nolimits_{j = 1}^{N_i^s} {x_{i,j}^s} + \sum\nolimits_{j = 1}^{N_i^q} {x_{i,j}^q})$. Then, we input $x_i$ into the encoder $g$, $\Xi$, and $f_{gr}$, i.e., ${f_{gr}}({\Xi ^{\rm{T}}}g({x_i}))$, yielding a vector with all elements greater than zero and matching the dimensionality of the generating representation. Then, each element is subject to the normalization operation, denoted as $\rm{Norm}(\cdot)$. As a result, the individual elements of the output vector, i.e., ${\rm{Norm}}(f_{gr})$, can be interpreted as the probabilities that each generating factor belongs to task $\tau_i$. 

Note that each task is associated with a subset of factors in $\Xi$ and can vary significantly from task to task. Meanwhile, the above calculation process of $\Xi$ and $f_{gr}$ may lead to degenerate solutions, e.g., the subset of generating factors for each task is the same. To address this issue, we propose a regularization term that consists of a $L_1$ norm and an entropy term, constraining the output of $f_{gr}$ to be sparse and diverse. By minimizing the $L_1$ norm, we make the output of $f_{gr}$ sparse, ensuring obtain subsets of generating factors only relevant to each single task. By maximizing the entropy term, we make the output of $f_{gr}$ diverse, preventing the acquisition of task-specific generating factors suffering degenerate solutions.
The regularization term is:
\begin{equation}
\label{eq:Lde}
\begin{array}{l}
\mathcal{L}_{\rm{DM}}(f_{gr})=\sum\limits_{i = 1}^{{N_{tr}}} {{{\left\| {{f_{gr}}({\Xi ^{\rm{T}}}g({x_i}))} \right\|}_1}} \\
 \quad\quad\quad\quad\quad\quad - {\rm{Entropy}}(\frac{{\sum\nolimits_j {{f_{gr}}{{({\Xi ^{\rm{T}}}g({x_i}))}_j}} }}{{\sum\nolimits_i {\sum\nolimits_j {{f_{gr}}{{({\Xi ^{\rm{T}}}g({x_i}))}_j}} } }})
\end{array}
\end{equation}
where ${{f_{gr}}{{({\Xi ^{\rm{T}}}g({x_i}))}_j}}$ represents the $j$-th element of the output of $f_{gr}$. Through Eq.\ref{eq:Lde}, we obtain accurate task-specific generating factors, thus eliminating task confounders.

By combining Eq.\ref{eq:Ldm} and Eq.\ref{eq:Lde}, we obtain the loss of the disentangling module which can be expressed as:
\begin{equation}
\label{eq:dis_loss}
    \mathcal{L}_{\rm{DM}}(f_{gr},\Xi) =  \lambda _{1}\cdot \mathcal{L}_{\rm{DM}}(\Xi) + \lambda _{2}\cdot \mathcal{L}_{\rm{DM}}(f_{gr})
\end{equation}
where $\lambda _{1}$ and $\lambda _{2}$ denote the loss weights of $\mathcal{L}_{\rm{DM}}(\Xi)$ and $\mathcal{L}_{\rm{DM}}(f_{gr})$, respectively. Through the above process with three constraints, i.e., correlation, sparsity, and diversity, we can accurately obtain all the generating factors and the task-specific generating factors without task confounders.

\subsection{Causal Module}
\label{sec:4.2}
In this module, we aim to ensure the causality of the generating factors obtained in the disentangling module. Following ~\cite{koyama2020invariance}, a model invariant to different distributions can learn causal correlations. Meanwhile, based on Theorem 9 described in \cite{DBLP:journals/corr/abs-1907-02893}, by enforcing invariance over multiple training datasets that exhibit distribution shifts, the task-specific models could only use task-related causal factors and assign zero weights to those non-causal generating factors. Therefore, the causal module is designed to facilitate causal learning by using this invariance, thereby ensuring the causality of the generating factors obtained by $\Xi$ and $f_{gr}$.

During the training phase of ML, the training data can be divided into multiple support sets and query sets. As they comprise different samples, they can be regarded as different data distributions with distributional shifts. Meanwhile, the learning process of meta-learning can be depicted as follows: First, for every $f_{\theta}$, optimizing Eq.\ref{eq:inner} can achieve an optimal $f_{\theta}^i$ and $\mathcal{L}(Y_i^s,X_i^s,f_{\theta}^i)$ on the support set. Next, altering the value of $f_{\theta}$ impacts the optimal $f_{\theta}^i$, we seek the optimal $f_{\theta}$ to obtain the optimal $f_{\theta}^i$ by optimizing $\frac{1}{N_{tr}}\sum_{i=1}^{N_{tr}}\mathcal{L}(Y_i^q,X_i^q,f_\theta ^i)$ on the query sets (Eq.\ref{eq:outer}). Thus, the bi-level optimization of Eq.\ref{eq:inner} and Eq.\ref{eq:outer} can be interpreted as achieving optimality across multiple datasets using the same $f_{\theta}$, and the causal factors are invariant on the support and query sets of the same task.

Based on the above illustration, we propose to utilize a bi-level optimization mechanism to learn $\Xi$ and $f_{gr}$ which is similar to Eq.\ref{eq:inner} and Eq.\ref{eq:outer}, thus ensuring causality. Specifically, for the first level, we learn $\Xi^{'}$ and ${f}_{gr}^{'}$ with the support sets through the following objectives:
\begin{equation}\label{eq:final1}
\begin{array}{l}
\qquad \qquad \qquad \left\{\begin{matrix}
\Xi^{'} \leftarrow \Xi - \alpha_{1} {\nabla _\Xi }\tilde{\mathcal{L}} \\
f_{gr}^{'} \leftarrow {f}_{gr} - \alpha_{2} {\nabla _{{f_{gr}}}}\tilde{\mathcal{L}}
\end{matrix}\right.\\[15pt]
s.t. \quad \tilde{\mathcal{L}}  =\frac{1}{N_{tr}} \sum_{i=1}^{N_{tr}} {\mathcal{L}(Y_i^{s},X_i^{s},\Xi,{{f}_{gr}})}  + {\mathcal{L}_{{\rm{DM}}}}(\Xi,{{f}_{gr}})\\[8pt]
\qquad \mathcal{L}(Y_i^{s},X_i^{s},\Xi, f_{gr}) = \frac{1}{{N_i^{s}}}\sum\nolimits_{j=1}^{N_i^{s}} y_{i,j}^{s} \log \mathrm{z}_{i,j}^{s} \\[8pt]
\qquad \mathrm {z}_{i,j}^{s}=h\{ {\rm{Norm[}}{f_{gr}}({\Xi ^{\rm{T}}}g({x_i}))] \odot [{\Xi ^{\rm{T}}}g(x_{i,j}^{s})]\}
\end{array}
\end{equation}
and for the second level, we learn ${\Xi}$ and ${{f}_{gr}}$ with the query sets through the following objectives:
\begin{equation}\label{eq:final2}
\begin{array}{l}
\qquad \qquad \qquad \left\{\begin{matrix}
\Xi \leftarrow \Xi - \alpha_{3} {\nabla _{\Xi} }\tilde{\mathcal{L}^{'}} \\
{f_{gr}} \leftarrow {{f}_{gr}} - \alpha_{4} {\nabla _{{f_{gr}}}}\tilde{\mathcal{L}^{'}}
\end{matrix}\right.\\[15pt]
s.t. \quad \tilde{\mathcal{L}^{'}}  =\frac{1}{N_{tr}} \sum_{i=1}^{N_{tr}} {\mathcal{L}(Y_i^{q},X_i^{q},\Xi^{'},f_{gr}^{'})}  + {\mathcal{L}_{{\rm{DM}}}}(\Xi^{'},f_{gr}^{'})\\[8pt]
\qquad \mathcal{L}(Y_i^{q},X_i^{q},\Xi^{'}, f_{gr}^{'}) = \frac{1}{{N_i^{q}}}\sum\nolimits_{j=1}^{N_i^{q}} y_{i,j}^{q} \log \mathrm{z}_{i,j}^{q} \\[8pt]
\qquad \mathrm {z}_{i,j}^{q}=h\{ {\rm{Norm[}}{f_{gr}}({{\Xi^{'}} ^{\rm{T}}}g({x_i}))] \odot [{{\Xi^{'}}^{\rm{T}}}g(x_{i,j}^{q})]\}
\end{array}
\end{equation}
where $\odot$ represents the element-wise multiplication operator between two vectors, i.e., the generating representation ${\Xi ^{\rm{T}}}g(x_{i,j}^{\cdot})$ and the weight ${\rm{Norm[}}{f_{gr}}({\Xi ^{\rm{T}}}g({x_i}))]$, while $\alpha_1$, $\alpha_2$, $\alpha_3$ and $\alpha_4$ are the learning rates. Note that both in Eq.\ref{eq:final1} and Eq.\ref{eq:final2}, the loss $\mathcal{L}(Y_i^{\cdot},X_i^{\cdot},\Xi ,{f_{gr}})$ is calculated using the generating representations with causal weights instead of feature representations, which restrict the features of the samples in $\tau_i$ to be associated only with task-specific causal factors.

In summary, the learning process of $\Xi$ and $f_{gr}$ can be regarded as enforcing invariance over the support sets and the query sets, and the bi-level optimization mechanism for $\Xi$ and $f_{gr}$ can ensure causality. Meanwhile, $\Xi$ and $f_{gr}$ are learned independently with the fixed meta-learning model $f_{\theta}$ in the middle training following modularity design, thus rendering the MetaCRL a plug-and-play learner.

\subsection{Overall Objective}
\label{sec:4.3}

In this subsection, we embed the above causal representation learning process into a meta-learning framework for joint optimization. The training process with MetaCRL in each batch is divided into two steps. In the first step, with $\Xi$ and $f_{gr}$ held fixed, we optimize the meta-learning model $f_{\theta}=h \circ g$. Specifically, the objective of the inner loop becomes:
\begin{equation}
\begin{array}{l}
    \qquad f^i_{\theta} \gets f_{\theta} -\alpha \nabla_{f_{\theta}}\tilde{\mathcal{L}} (Y_i^s,X_i^s,f_{\theta}) \\[5pt]
    s.t. \quad \tilde{\mathcal{L}}(Y_i^s,X_i^s,f_{\theta} )=\frac{1}{N_i^s} \sum_{j=1}^{N_i^s}y_{i,j}^s\log \mathrm {z}_{i,j}^{s}
\end{array}
\end{equation}
where $\mathrm {z}_{i,j}^{s}$ is calculated the same as Eq.\ref{eq:final1}. Subsequently, the objective of the outer loop mentioned in Eq.\ref{eq:outer} becomes:
\begin{equation}
\begin{array}{l}
    \quad f_{\theta} \gets f_{\theta}-\beta \nabla_{f_{\theta}}\frac{1}{N_{tr}}\sum_{i=1}^{N_{tr}}\tilde{\mathcal{L}}(Y_i^q,X_i^q,f_\theta ^i)  \\[5pt]
    s.t. \quad \tilde{\mathcal{L}}(Y_i^q,X_i^q,f_\theta ^i)= \frac{1}{N_i^q} \sum_{j=1}^{N_i^q}y_{i,j}^q\log \mathrm {z}_{i,j}^{q}
\end{array}
\end{equation}
where $\mathrm {z}_{i,j}^{q}$ is calculated as mentioned in Eq.\ref{eq:final2}. Next, in the second step, with the meta-learning model $f_{\theta}$ held fixed, we optimize $\Xi$ and $f_{gr}$ as mentioned in Eq.\ref{eq:final1} and Eq.\ref{eq:final2}.

By incorporating the causal invariant-based optimization mechanism and the additional regularization term, we can effectively eliminate task confounders that lead to model degradation and improve generalization capability.



\section{Experiments}
\label{sec:5}

In this section, we first evaluate MetaCRL on various scenarios, including sinusoid regression, image classification, drug activity prediction, and pose prediction in Subsections \ref{sec:5.1}-\ref{sec:5.4}, respectively. Next, we conduct ablation studies and visualization in Subsections \ref{sec:5.5} and \ref{sec:5.6}. Considering that MetaCRL is a plug-and-play method, we assess its performance on several meta-learning models, e.g., MAML~\cite{finn2017model}, ANIL~\cite{raghu2019rapid}, MetaSGD~\cite{li2017meta}, and T-NET~\cite{lee2018gradient}, and multiple causal-based baselines, e.g., IFSL ~\cite{yue2020interventional}, Meta-Trans \cite{bengio2019meta}, Meta-Aug \cite{meta-aug}, and MR-MAML ~\cite{yin2019meta}, to demonstrate its compatibility. Considering that MetaCRL addresses the ``Task Confounder'' problem to enhance generalization, we also compare it with the plug-and-play generalization baselines that are most relevant to our method, i.e., MetaMix~\cite{yao2021improving} and Dropout-Bins~\cite{jiang2022role}. We delay all the details of datasets, baselines, implementation details, and additional experimental results in Appendices C-F, respectively.

\begin{table}[t]
\begin{center}
\resizebox{0.8\linewidth}{!}{
\begin{tabular}{l|c|c}
\toprule
\textbf{Model} & \textbf{5-shot} & \textbf{10-shot}\\
\midrule
IFSL  &  0.592 $\pm$ 0.141 & 0.178 $\pm$ 0.040 \\
Meta-Trans & 0.577 $\pm$ 0.123 & 0.140 $\pm$ 0.024 \\
Meta-Aug & 0.531 $\pm$ 0.118 & 0.103 $\pm$ 0.031 \\
MR-MAML   &  0.581 $\pm$ 0.110 & 0.104 $\pm$ 0.029 \\
\midrule
MAML  & 0.593 $\pm$ 0.120 & 0.166 $\pm$ 0.061 \\
MAML + MetaMix  & 0.476 $\pm$ 0.109 & 0.085 $\pm$ 0.024 \\
MAML + Dropout-Bins  & 0.452 $\pm$ 0.081 & 0.062 $\pm$ 0.017 \\
\rowcolor{orange!10}\textbf{MAML + Ours}  & \textbf{0.440 $\pm$ 0.079} & \textbf{0.054 $\pm$ 0.018} \\
\midrule
ANIL & 0.541 $\pm$ 0.118 & 0.103 $\pm$ 0.032 \\
ANIL + MetaMix & 0.514 $\pm$ 0.106 & 0.083 $\pm$ 0.022 \\
ANIL + Dropout-Bins  & 0.487 $\pm$ 0.110 & 0.088 $\pm$ 0.025 \\
\rowcolor{orange!10}\textbf{ANIL + Ours}  & \textbf{0.468 $\pm$ 0.094} & \textbf{0.081 $\pm$ 0.019} \\
\midrule
MetaSGD  & 0.577 $\pm$ 0.126 & 0.152 $\pm$ 0.044 \\
MetaSGD + MetaMix  & 0.468 $\pm$ 0.118 & 0.072 $\pm$ 0.023 \\
MetaSGD + Dropout-Bins  & 0.435 $\pm$ 0.089 & 0.040 $\pm$ 0.011 \\
\rowcolor{orange!10}\textbf{MetaSGD + Ours}  & \textbf{0.408 $\pm$ 0.071} & \textbf{0.038 $\pm$ 0.010} \\
\midrule
T-NET  & 0.564 $\pm$ 0.128 & 0.111 $\pm$ 0.042 \\
T-NET + MetaMix  & 0.498 $\pm$ 0.113 & 0.094 $\pm$ 0.025\\
T-NET + Dropout-Bins  & 0.470 $\pm$ 0.091 & 0.077 $\pm$ 0.028 \\
\rowcolor{orange!10}\textbf{T-NET + Ours}  & \textbf{0.462 $\pm$ 0.078} & \textbf{0.071 $\pm$ 0.019} \\
\bottomrule
\hline
\end{tabular}}
\end{center}
\vspace{-0.1in}
\caption{Performance (MSE) comparison on the sinusoid regression problem. ``+ours'' means integrating MetaCRL into the existing methods, and the best results are highlighted in \textbf{bold}.}
\label{tab:2}
\end{table}

\begin{table}[t]
\begin{center}
\resizebox{0.9\linewidth}{!}{
\begin{tabular}{l|c|c|c}
\toprule
\textbf{Model} & \textbf{Omniglot} & \textbf{miniImagenet} & \textbf{TC}\\
\midrule
IFSL  & 88.51 $\pm$ 0.49 & 36.21 $\pm$ 1.62 & $\setminus $ \\
Meta-Trans & 87.39 $\pm$ 0.51 & 35.19 $\pm$ 1.58 & $\setminus $ \\
Meta-Aug & 89.77 $\pm$ 0.62 & 34.76 $\pm$ 1.52 & $\setminus $ \\
MR-MAML & 89.28 $\pm$ 0.59 & 35.01 $\pm$ 1.60 & $\setminus $ \\
\midrule
MAML  & 87.15 $\pm$ 0.61 & 33.16 $\pm$ 1.70 & 0.00 \\
MAML + MetaMix  & 91.97 $\pm$ 0.51 & 38.97 $\pm$ 1.81 & +0.42 \\
MAML + Dropout-Bins  & 92.89 $\pm$ 0.46 & 39.66 $\pm$ 1.74 & -0.14 \\
\rowcolor{orange!10}\textbf{MAML + Ours}  & \textbf{93.00 $\pm$ 0.42} & \textbf{41.55 $\pm$ 1.76} & \textbf{+4.12} \\
\midrule
ANIL   & 89.17 $\pm$ 0.56 & 34.96 $\pm$ 1.71 & 0.00 \\
ANIL + MetaMix  & 92.88 $\pm$ 0.51 & 37.82 $\pm$ 1.75 & -0.10 \\
ANIL + Dropout-Bins  & 92.82 $\pm$ 0.49 & 38.09 $\pm$ 1.76 & +0.97 \\
\rowcolor{orange!10}\textbf{ANIL + Ours}  & \textbf{92.91 $\pm$ 0.52} & \textbf{38.55 $\pm$ 1.81} & \textbf{+3.56} \\
\midrule
MetaSGD  & 87.81 $\pm$ 0.61 & 33.97 $\pm$ 0.92 & 0.00 \\
MetaSGD + MetaMix  & 93.44 $\pm$ 0.45 & 40.28 $\pm$ 0.96 & +0.05 \\
MetaSGD + Dropout-Bins  & 93.93 $\pm$ 0.40 & 40.31 $\pm$ 0.96 & +1.08 \\
\rowcolor{orange!10}\textbf{MetaSGD + Ours}  & \textbf{94.12 $\pm$ 0.43} & \textbf{41.22 $\pm$ 0.93} & \textbf{+6.19} \\
\midrule
T-NET  & 87.66 $\pm$ 0.59 & 33.69 $\pm$ 1.72 & 0.00 \\
T-NET + MetaMix  & 93.16 $\pm$ 0.48 & 39.18 $\pm$ 1.73 & +0.28 \\
T-NET + Dropout-Bins  & 93.54 $\pm$ 0.49 & 39.06 $\pm$ 1.72 & +1.03 \\
\rowcolor{orange!10}\textbf{T-NET + Ours}  & \textbf{93.81 $\pm$ 0.52} & \textbf{40.08 $\pm$ 1.74} & \textbf{+4.65} \\
\bottomrule
\hline
\end{tabular}}
\end{center}
\vspace{-0.1in}
\caption{Performance (accuracy $\pm $ 95\% confidence interval) on (20-way 1-shot) Omniglot and (5-way 1-shot) miniImagenet. The ``+'' and ``-'' indicate the performance changes, and the ``$\setminus $'' denotes that the result is not reported. See Appendix F for full results. }
\label{tab:3}
\end{table}

\begin{table*}
\begin{center}
\resizebox{0.85\linewidth}{!}{
\begin{tabular}{l|ccc|ccc|ccc|ccc}
\toprule
\multirow{2}{*}{\textbf{Model}}
    & \multicolumn{3}{c|}{\textbf{Group 1}} & \multicolumn{3}{c|}{\textbf{Group 2}} & \multicolumn{3}{c|}{\textbf{Group 3}} & \multicolumn{3}{c}{\textbf{Group 4}} \\ 
    & \textbf{Mean} & \textbf{Med.} & \textbf{$>$ 0.3} & \textbf{Mean} & \textbf{Med.} & \textbf{$>$ 0.3} & \textbf{Mean} & \textbf{Med.} & \textbf{$>$ 0.3} & \textbf{Mean} & \textbf{Med.} & \textbf{$>$ 0.3} \\
\midrule
MAML  & 0.371 &  0.315 & 52 & 0.321 &  0.254 & 43 & 0.318 & 0.239 & 44 & 0.348 & 0.281 & 47 \\
MAML + Dropout-Bins  & 0.410 & 0.376 & 60 & 0.355 & 0.257 & 48 & 0.320 & 0.275 & 46 & 0.370 & 0.337 & 56 \\
\rowcolor{orange!10}MAML + Ours  & \textbf{0.413} & \textbf{0.378} & \textbf{61} & \textbf{0.360} & \textbf{0.261} & \textbf{50} & \textbf{0.334} & \textbf{0.282} & \textbf{51} & \textbf{0.375} & \textbf{0.341} & \textbf{59} \\
\midrule
ANIL  & 0.355 & 0.296 & 50 & 0.318 & \textbf{0.297} & \textbf{49} & 0.304 & 0.247 & 46 & 0.338 & 0.301 & 50  \\
ANIL + MetaMix  & 0.347 & 0.292 & 49 & 0.302 & 0.258 & 45 & 0.301 & 0.282 & 47 & 0.348 & 0.303 & 51  \\
ANIL + Dropout-Bins  & 0.394 & 0.321 & 53 & 0.338 & 0.271 & 48 & \textbf{0.312} & 0.284 & 46 & 0.368 & 0.297 & 50 \\
\rowcolor{orange!10}ANIL + Ours  & \textbf{0.401} & \textbf{0.339} & \textbf{57} & \textbf{0.341} & 0.277 & \textbf{49} & \textbf{0.312} & \textbf{0.291} & \textbf{48} & \textbf{0.371} & \textbf{0.305} & \textbf{53}  \\
\bottomrule
\end{tabular}}
\end{center}
\vspace{-0.1in}
\caption{Performance comparison on drug activity prediction. ``Mean'', ``Med.'', and ``$> 0.3$" are the mean, the median value of $R^2$, and the number of analyzes for $R^2> 0.3$. The best results are highlighted in \textbf{bold}.}
\label{tab:4}
\end{table*}

\begin{table}
\begin{center}
\resizebox{0.8\linewidth}{!}{
\begin{tabular}{l|c|c}
\toprule
\textbf{Model} & \textbf{10-shot} & \textbf{15-shot}\\
\midrule
MAML  & 3.113 $\pm$ 0.241  & 2.496 $\pm$ 0.182 \\
MAML + MetaMix  & 2.429 $\pm$ 0.198 & 1.987 $\pm$ 0.151 \\
MAML + Dropout-Bins  & 2.396 $\pm$ 0.209 & 1.961 $\pm$ 0.134 \\
\rowcolor{orange!10}\textbf{MAML + Ours}  & \textbf{2.355 $\pm$ 0.200} & \textbf{1.931 $\pm$ 0.134} \\
\midrule
MetaSGD  & 2.811 $\pm$ 0.239 & 2.017 $\pm$ 0.182 \\
MetaSGD + MetaMix  & 2.388 $\pm$ 0.204 & 1.952 $\pm$ 0.134 \\
MetaSGD + Dropout-Bins  & 2.369 $\pm$ 0.217  & 1.927 $\pm$ 0.120 \\
\rowcolor{orange!10}\textbf{MetaSGD + Ours}  & \textbf{2.362 $\pm$ 0.196} & \textbf{1.920 $\pm$ 0.191} \\
\midrule
T-NET & 2.841 $\pm$ 0.177 & 2.712 $\pm$ 0.225 \\
T-NET + MetaMix  & 2.562 $\pm$ 0.280 & 2.410 $\pm$ 0.192 \\
T-NET + Dropout-Bins  & 2.487 $\pm$ 0.212 & 2.402 $\pm$ 0.178 \\
\rowcolor{orange!10}\textbf{T-NET + Ours}  & \textbf{2.481 $\pm$ 0.274}  & \textbf{2.400 $\pm$ 0.171} \\
\bottomrule
\hline
\end{tabular}}
\end{center}
\vspace{-0.1in}
\caption{Performance (MSE $\pm $ 95\% confidence interval) comparison on pose prediction. More results are provided in Appendix F.}
\label{tab:5}
\end{table}

\subsection{Sinusoid Regression}
\label{sec:5.1}

Firstly, we evaluate the performance of our MetaCRL on sinusoid regression. Following ~\cite{jiang2022role}, we conduct 480 tasks and the data for each task is generated in the form of $A\sin w \cdot x + b + \epsilon $, where $A \in \left [ 0.1, 5.0 \right ] $, $w \in \left [ 0.5, 2.0 \right ]$, and $b \in \left [ 0,2\pi \right ] $. We add Gaussian observation noise with $\mu = 0$ and $\epsilon = 0.3$ to each data point sampled from the target task. In this experiment, we set $\lambda_{1}$ and $\lambda_{2}$ to 0.4 and 0.2. We use the Mean Squared Error (MSE) as the evaluation metric.

The results are shown in Table \ref{tab:2}. Compared to the plug-and-play baselines, MetaCRL achieves improvements with an average MSE reduction of 0.034 and 0.013, respectively. MetaCRL also demonstrates significant improvements across all the meta-learning base models, with an MSE reduction of over 0.1. Compared to the causal-based baselines, adding MetaCRL to any meta-learning model can always achieve better performance. As expected, MetaCRL exhibits significant enhancements, showcasing its high compatibility.

\subsection{Image Classification}
\label{sec:5.2}

Next, we conduct experiments on image classification, utilizing two benchmark datasets, i.e., miniImagenet and Omniglot. These two datasets contain 600 and 1623 tasks, respectively. We also introduce a specialized dataset called ``TC'', which comprises 50 groups of tasks (300 tasks in total) identified as being affected by task confounders, i.e., tasks with negative knowledge transfer as mentioned in Subsection \ref{sec:3.2}. More details are provided in Appendix C. In this experiment, we set $\lambda_{1}$ and $\lambda_{2}$ to 0.5 and 0.35, respectively. The evaluation metric employed here is the average accuracy.

The results are shown in Table \ref{tab:3}. MetaCRL consistently surpasses the SOTA baselines across all datasets, indicating that it can achieve better generalization improvements than the baselines do without the need for task-specific or general-label space augmentation that the baselines need. Notably, on the ``TC'' dataset, MetaCRL outperforms the baselines by a significant margin, which demonstrates a unique advantage of MetaCRL in handling task confounders. In summary, MetaCRL continues to exhibit remarkable performance and adeptly eliminates task confounders.

\subsection{Drug Activity Prediction}
\label{sec:5.3}

We also evaluate MetaCRL on drug activity prediction. pQSAR~\cite{martin2019all} is a dataset designed to forecast the activity of compounds on specific target proteins, encompassing a total of 4276 tasks. We adopt the same settings as~\cite{yao2021improving} and divide the tasks into four groups. In this experiment, $\lambda_{1}$ and $\lambda_{2}$ are both set to 0.3, and the evaluation metric is the squared Pearson correlation coefficient ($R^2$), reflecting the correlation between predictions and the actual values for each task. We record both the mean and median $R^2$ values, along with the count of $R^2$ values exceeding 0.3, which stands as a reliable indicator in pharmacology.

The results are shown in Table \ref{tab:4}. MetaCRL attains performance levels akin to the SOTA baselines across all four groups of data. Notably, we achieve a noteworthy enhancement of 3 in the reliability index $R^2 > 0.3$. The achievement of this scenario underscores the effectiveness of our MetaCRL across disparate domains and the pervasive influence of task confounders. See Appendix F for full results.

\begin{figure}[t]
    \centering
    \begin{subfigure}{0.24\columnwidth}
        \centering
        \includegraphics[width=\linewidth]{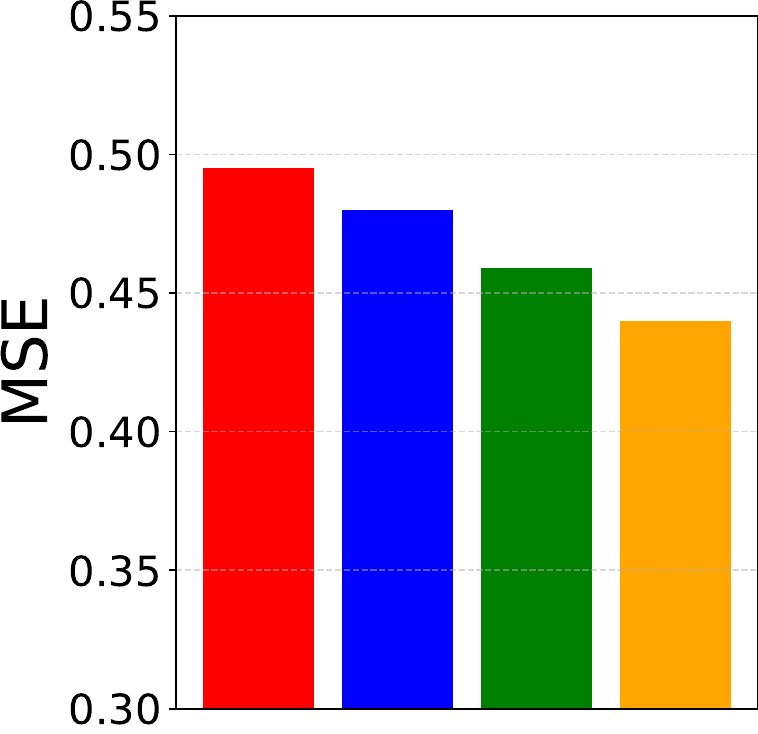}
        \vspace{-0.2in}
        \caption{}
        \label{fig:subfig1}
        \vspace{-0.1in}
    \end{subfigure}
    \hfill
    \begin{subfigure}{0.23\columnwidth}
        \centering
        \includegraphics[width=\linewidth]{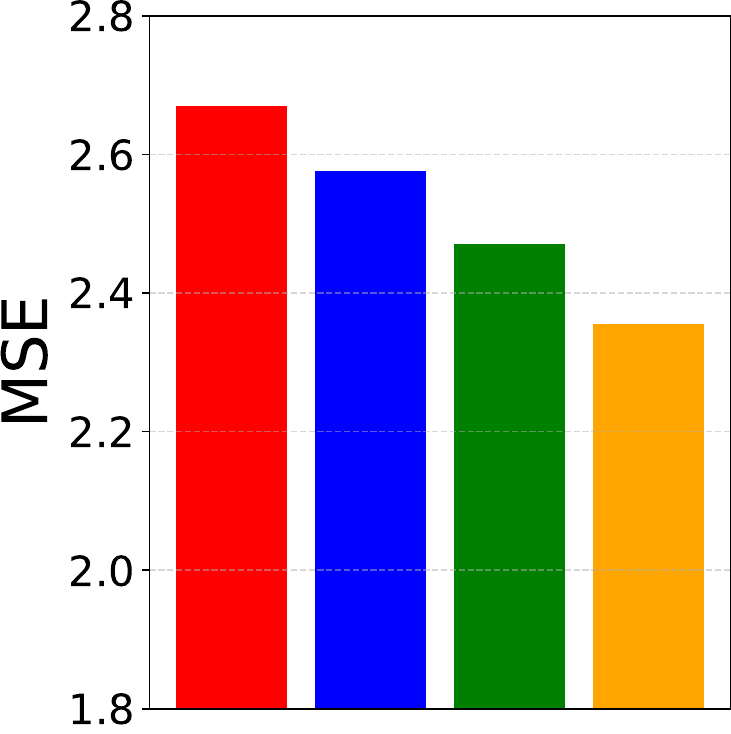}
        \vspace{-0.2in}
        \caption{}
        \label{fig:subfig2}
        \vspace{-0.1in}
    \end{subfigure}
    \hfill
    \begin{subfigure}{0.23\columnwidth}
        \centering
        \includegraphics[width=\linewidth]{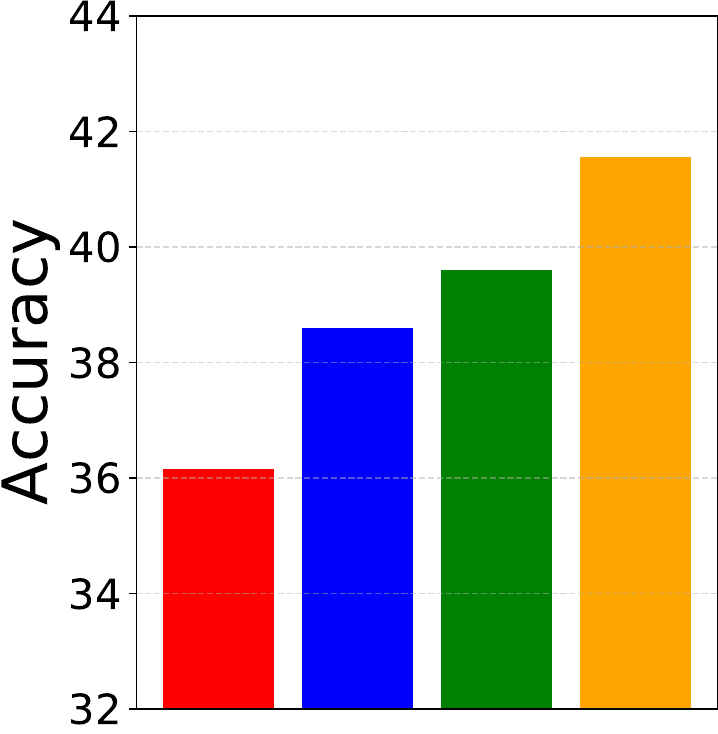}
        \vspace{-0.2in}
        \caption{}
        \label{fig:subfig3}
        \vspace{-0.1in}
    \end{subfigure}
    \hfill
    \begin{subfigure}{0.23\columnwidth}
        \centering
        \includegraphics[width=\linewidth]{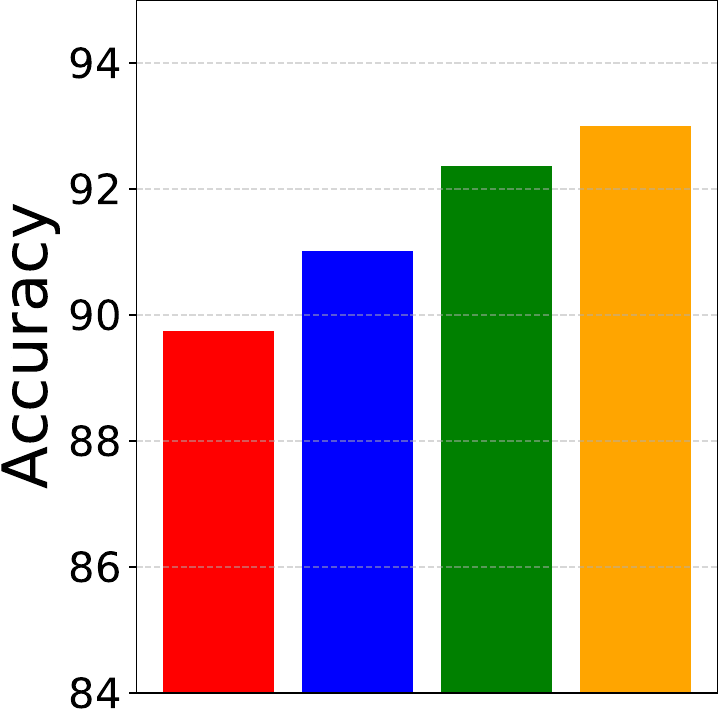}
        \vspace{-0.2in}
        \caption{}
        \label{fig:subfig4}
        \vspace{-0.1in}
    \end{subfigure}
\caption{Ablation study, including (a) sinusoid regression, (b) pose prediction, (c) 5-way 1-shot miniImagenet, and (d) 20-way 1-shot Omniglot. The backbone is MAML. The \textcolor{red}{red}, \textcolor{blue}{blue}, \textcolor{green}{green}, and \textcolor{orange}{orange} bars represent the results of MetaCRL-$\mathcal{L}_{\rm{DM}}(f_{gr},\Xi)$, MetaCRL-$\mathcal{L}_{\rm{DM}}(\Xi)$, MetaCRL-$\mathcal{L}_{\rm{DM}}(f_{gr})$, and MetaCRL.}
\label{fig:3}
\end{figure}

\subsection{Pose Prediction}
\label{sec:5.4}

Lastly, we undertake the fourth benchmark, focusing on pose prediction. This evaluation is constructed using the Pascal 3D dataset ~\cite{xiang2014beyond}. We randomly select 50 objects for meta-training and 15 additional objects for meta-testing. In this experiment, the values of $\lambda_{1}$ and $\lambda_{2}$ are set to 0.3 and 0.2, while the evaluation metric employed here is MSE.

The results are shown in Table \ref{tab:5}. MetaCRL achieves the best performance. Notably, drawing insights from the findings presented in ~\cite{yao2021improving}, we posit that augmenting the dataset could yield more effective results in this scenario, potentially outperforming the reliance solely on meta-regularization techniques. MetaCRL incorporates regularization terms instead of data augmentation and still manages to achieve enhanced performance, thereby affirming its efficacy.

\subsection{Ablation Study}
\label{sec:5.5}

We conduct ablation studies to explore the impact of different regularization terms, that is $\mathcal{L}_{\rm{DM}}(\Xi) $, $\mathcal{L}_{\rm{DM}}(f_{gr})$, and their combination $\mathcal{L}_{\rm{DM}}(f_{gr},\Xi)$ in Eq.\ref{eq:dis_loss}. We select both classification and regression scenarios, including four benchmark datasets. Figure \ref{fig:3} shows the results that $\mathcal{L}_{\rm{DM}}(\Xi) $ and $\mathcal{L}_{\rm{DM}}(f_{gr})$ promote the model in all datasets, and the improvement is the largest when combined. Moreover, despite eliminating the regularization terms, MetaCRL still significantly outperforms the base models, illustrating the effectiveness of the causal module. We also construct ablation studies targeting the accuracy of extracting task-specific causal factors and model efficiency (See Appendix F for details).

\begin{figure}
    \begin{minipage}[t]{0.47\columnwidth}
    \vspace{-0.2in}
        \centering
        \includegraphics[width=\linewidth]{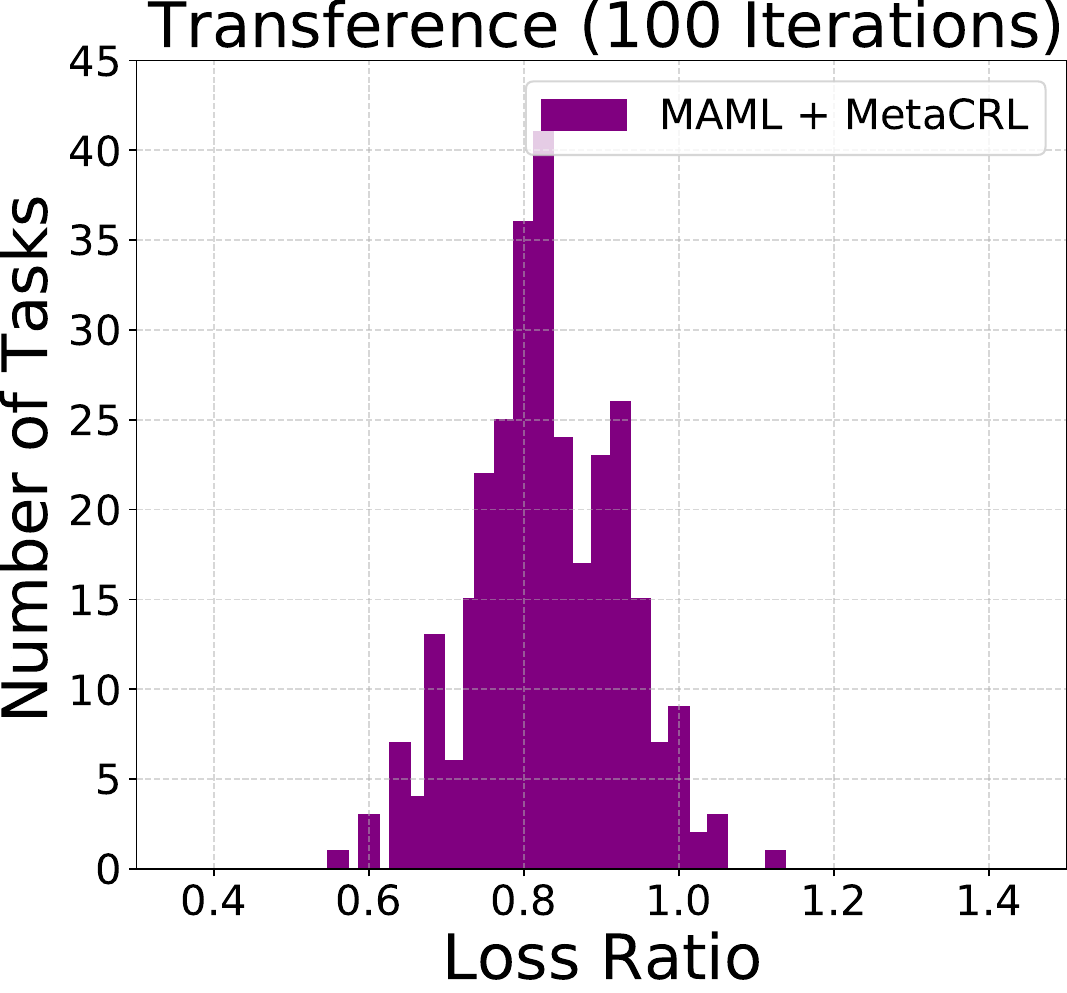}
        \vspace{-0.25in}
        \caption{Knowledge transference after using MetaCRL.}
        \label{fig:batchsize}
    \end{minipage}
    \hfill
    \begin{minipage}[t]{0.5\columnwidth}
    \vspace{-0.2in}
        \centering
        \includegraphics[width=\linewidth]{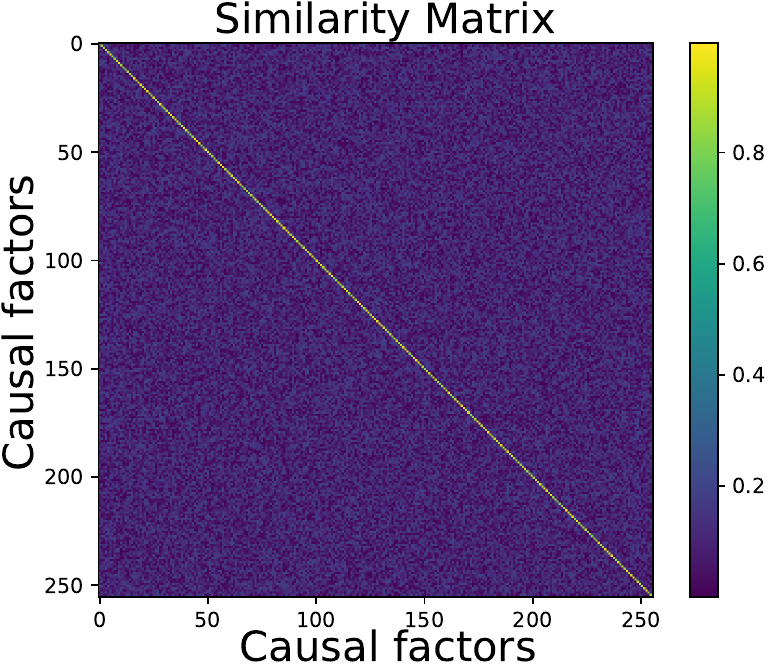}
        \vspace{-0.25in}
        \caption{Visualization of the similarity between causal factors.}
        \label{fig:similar}
    \end{minipage}
\end{figure}

\subsection{Visualization}
\label{sec:5.6}
To better evaluate the effect of MetaCRL, we visualize (i) knowledge transfer after using MetaCRL; and (ii) the similarity between causal factors. The former evaluates MetaCRL's efficacy in ensuring causality and avoiding negative knowledge transfer caused by task confounders, which use the same settings as in Subsection \ref{sec:3.2}. The latter assesses the decoupling of causal factors using cosine similarity. Figures \ref{fig:batchsize} and \ref{fig:similar} show visualizations for these two aspects, respectively. Figure \ref{fig:batchsize} shows that there are almost no training tasks that lead to negative knowledge transfer with fewer iterations than Figure \ref{fig:intro}, which indicates that MetaCRL effectively eliminates task confounders. Figure \ref{fig:similar} shows that the similarity scores between different causal factors are very low, illustrating that the disentangling module successfully decouples causal factors. More details are provided in Appendix F.



\section{Conclusion}
\label{sec:6}

In this paper, we discover a valuable problem called ``Task Confounder'', and propose a novel method called MetaCRL to address its unique challenges. We begin by analyzing a counterintuitive negative knowledge transfer phenomenon with SCM, revealing spurious correlations between causal factors of the training tasks and the label space, i.e., ``Task Confounder". Then, we propose MetaCRL, which consists of two modules: (i) a disentangling module that acquires generating factors and eliminates task confounders; and (ii) a causal module that ensures causality of the obtained generating factors. It is a plug-and-play causal representation learner that can be applied to any meta-learning baseline. Extensive experiments demonstrate the effectiveness and robustness of MetaCRL. Our work uncovers a novel and significant issue in ML, providing valuable insights for future research.

\section*{Acknowledgements}
The authors would like to thank the anonymous reviewers for their valuable comments. This work is supported in part by the Postdoctoral Fellowship Program of CPSF No. GZB20230790, the China Postdoctoral Science Foundation No. 2023M743639, and the Special Research Assistant Fund, Chinese Academy of Sciences No. E3YD590101. The Appendix is provided in \url{https://arxiv.org/abs/2312.05771}.

\section*{Contribution Statement}
Jingyao Wang and Yi Ren made equal contributions. All the authors participated in designing research, performing research, analyzing data, and writing the paper.

\bibliographystyle{named}
\bibliography{ijcai24}

\newpage
\appendix
The appendix provides supplementary information and additional details that support the primary discoveries and methodologies proposed in this paper. It is organized into several sections: Appendix A contains the proof of Theorem 1, while Appendix B encompasses the pseudo-code and the pipeline of the meta-learning process using MetaCRL. Appendices C and D provide details for all datasets and baselines mentioned in the main text. Appendix E presents the implementation and architecture of our method, aiding in the faithful reproduction of our work. Furthermore, Appendix F showcases additional experiments and full results that were omitted in the main paper due to page limitations.

\section{Proof of Theorem 1}

In this section, we will prove Theorem 1, which is used for causal analysis and motivation.

\subsection{Problem Definition}

In the analysis of this theorem, we consider a simple scenario of two binary classification tasks, denoted as $\tau_i$ and $\tau_j$. We use $Y_i$ and $Y_j$ to respectively represent the label variables of tasks $\tau_i$ and $\tau_j$, while $X_i$ and $X_j$ represent the sample variables of the two tasks. Since the tasks are binary classification tasks, $Y_i$ and $Y_j$ can be considered as variables belonging to the set of task labels $\left\{ { \pm 1} \right\}$. Note that any multi-classification task is a combination of binary tasks (the current and other classes). In this proof, we choose binary tasks to prove task confounder more simply and directly.

We assume that these labels are drawn from two different probabilities, and the sampling probabilities of label values are balanced, i.e., $P(Y=1)=P(Y=-1)=0.5$. Our conclusions also hold for imbalanced distributions.

Given the set of causal factors for the entire world $\mathrm {A}^{w}$, the training set reflects a part of the world with the causal factors $\mathrm {A}^{tr} \in \mathrm {A}^{w}$. Since $\mathrm {A}^{tr}$ is unknown, we use a Gaussian distribution to model $\mathrm {A}^{w}$, and the probability of the causal factor reflects its likelihood of belonging to $\mathrm {A}^{tr}$. 

Thus, for task $\tau_i$ and $\tau_j$, we consider two non-overlapping factors representing knowledge in $N_z$ dimensions, denoted as ${\rm A}^i$ and ${\rm A}^j$, to address these two tasks. Both factors are assumed to be drawn from Gaussian distributions:

\begin{equation}
\begin{array}{l}
    {\rm A}^i\sim \mathcal{N}(Y_i\cdot \mu_i,\sigma_i^2 I)\\[8pt]
    {\rm A}^j\sim \mathcal{N}(Y_j\cdot \mu_j,\sigma_j^2 I)
\end{array}
\end{equation}
where $\mu_i,\mu_j\in \mathbb{R}^{N_z} $ denote the mean vectors, while $\sigma_i^2$ and $\sigma_j^2$ denote the covariance vectors. 

In this paper, we focus on the spurious correlations between tasks caused by task confounders. For the sake of simplicity, we define $p_{tc}$ to represent the varying correlations resulting from different task confounders across different batches. Hence, $P(Y_i=Y_j)=p_{tc}$ and $P(Y_i\neq Y_j)=1-p_{tc}$. Further, we can formulate the probability table as follows:

\begin{table}[ht]
\begin{center}
\begin{tabular}{c|cc}
\toprule
 & $Y_i=1$ & $Y_i=-1$\\
\midrule
$Y_j=1$ & $p_{tc}$ & 1-$p_{tc}$ \\
$Y_j=-1$ & 1-$p_{tc}$ & $p_{tc}$ \\
\bottomrule
\hline
\end{tabular}
\end{center}
\end{table}

when $p_{tc}$ equals 0.5, it indicates that under this circumstance, the two tasks $\tau_i$ and $\tau_j$ are correlated within these environments. In this case, our objective is to learn two linear models, $P(Y_i|{\rm A}^i,{\rm A}^j)$ and $P(Y_j|{\rm A}^i,{\rm A}^j)$, guided by the following theorem:

\begin{theorem} \label{theorem1}
If $p_{tc}\ne 0.5$ (the correlation between $Y_i$ and $Y_j$ is not equal to 0.5), the optimal classifier has non-zero weights for non-causal factors for each task. If $p_{tc}= 0.5$ and the number of training samples is limited, the optimal classifier also has non-zero weights for non-causal factors for each task.
\end{theorem}

Assuming that there is no traditional spurious correlation among factor labels in single-task learning, the optimal classifier will consider only causal factors as features while assigning zero weight to non-causal factors.

\subsection{Proof}

With the presence of $p_{tc}$, when training a single model using two tasks, the optimal classifier for the target task will incorporate causal features from the other task that are non-causal factors. To substantiate this, we assume the implementation of a Bayesian classifier. Taking task $\tau_i$ as an illustration, we proceed to deduce the optimal Bayesian classifier as follows:
\begin{equation}\label{eq:P}
\begin{aligned}
    P(Y_i|{\rm A}^i,{\rm A}^j) &=\frac{P(Y_i,{\rm A}^i,{\rm A}^j)}{P({\rm A}^i,{\rm A}^j)} \\
    &=\frac{P(Y_i,{\rm A}^i,{\rm A}^j)}{ {\textstyle \sum_{Y_i\in \left \{ -1,1 \right \}  }} P(Y_i,{\rm A}^i,{\rm A}^j)} 
\end{aligned}
\end{equation}
where the probability of $P(Y_i,{\rm A}^i,{\rm A}^j)$ could be written as:
\begin{equation}
\begin{array}{l}
     P(Y_i,{\rm A}^i,{\rm A}^j) = P(Y_i,{\rm A}^i)\cdot P({\rm A}^j|Y_i,{\rm A}^i)\\[8pt]
        = P(Y_i,{\rm A}^i)\cdot P({\rm A}^j|Y_i)\\[8pt]
        = P(Y_i,{\rm A}^i)\cdot\sum_{Y_j\in \left \{ -1,1 \right \} }P({\rm A}^j,Y_j|Y_i)\\[8pt]
        = P(Y_i)P({\rm A}^i|Y_i)\cdot\sum_{Y_j\in \left \{ -1,1 \right \} }P({\rm A}^j|Y_j)P(Y_j|Y_i)
\end{array}
\end{equation}
Consider both ${\rm A}^i$ and ${\rm A}^j$ are assumed to be drawn from Gaussian distributions, and $P(Y_{i/j},{\rm A}^i,{\rm A}^j)=sigmoid(\frac{\mu _i}{\sigma^2_i}{\rm A}^i+\frac{\mu _j}{\sigma^2_j}{\rm A}^j)$ where $\frac{\mu _i}{\sigma^2_i}$ and $\frac{\mu _j}{\sigma^2_j}$ are the regression vectors for the optimal Bayesian classifier, then we have:
\begin{equation}\label{eq:P_range}
\begin{array}{l}
     P(Y_i,{\rm A}^i,{\rm A}^j) = P(Y_i,{\rm A}^i)\cdot P({\rm A}^j|Y_i,{\rm A}^i)\\[8pt]
     = P(Y_i)P({\rm A}^i|Y_i)\cdot\sum_{Y_j\in \left \{ -1,1 \right \} }P({\rm A}^j|Y_j)P(Y_j|Y_i) \\[8pt]
     \propto e^{Y_i\cdot \frac{\mu _i}{\sigma^2_i}{\rm A}^i}(p_{tc}e^{Y_i\cdot \frac{\mu _j}{\sigma^2_j}{\rm A}^j}+(1-p_{tc})e^{-Y_i\cdot \frac{\mu _j}{\sigma^2_j}{\rm A}^j}) \\[8pt]
     =p_{tc}e^{Y_i\cdot (\frac{\mu _i}{\sigma^2_i}{\rm A}^i+\frac{\mu _j}{\sigma^2_j}{\rm A}^j)}+(1-p_{tc})e^{Y_i\cdot (\frac{\mu _i}{\sigma^2_i}{\rm A}^i-\frac{\mu _j}{\sigma^2_j}{\rm A}^j)}
\end{array}
\end{equation}
Let:
\begin{equation}\label{eq:beta}
\begin{array}{l}
     \beta^{+}=\frac{\mu _i}{\sigma^2_i}{\rm A}^i+\frac{\mu _j}{\sigma^2_j}{\rm A}^j\\[8pt]
     \beta^{-}=\frac{\mu _i}{\sigma^2_i}{\rm A}^i-\frac{\mu _j}{\sigma^2_j}{\rm A}^j
\end{array}
\end{equation}
Then, by putting Eq.\ref{eq:P_range} back to Eq.\ref{eq:P}, we have:
\begin{equation}\label{eq:P_final}
\begin{aligned}
    P(Y_i|{\rm A}^i,{\rm A}^j) =\frac{1}{1+\frac{p_{tc}e^{Y_i\cdot \beta ^{+}}+(1-p_{tc})e^{Y_i\cdot \beta^- }}{p_{tc}e^{-Y_i\cdot \beta ^{+}}+(1-p_{tc})e^{-Y_i\cdot \beta^- }} } 
\end{aligned}
\end{equation}
\textbf{When $p_{tc}=0.5$}:
\begin{equation}
\begin{aligned}
    P(Y_i|{\rm A}^i,{\rm A}^j) =\frac{1}{1+e^{Y_i\cdot (\beta^+ +\beta^-)}}
\end{aligned}
\end{equation}
Combining Eq.\ref{eq:beta}, then we have:
\begin{equation}
\begin{aligned}
    P(Y_i|{\rm A}^i,{\rm A}^j) =\frac{1}{1+e^{2Y_i\cdot (\frac{\mu _i}{\sigma^2_i}{\rm A}^i)}}
\end{aligned}
\end{equation}
In this way, the optimal classifier for task $\tau_i$ only utilizes its factor ${\rm A}^i$ and assigns zero weights to the non-causal factor ${\rm A}^j$ which belongs to task $\tau_j$. That is, \emph{if $p_{tc}= 0.5$, the optimal classifier also has non-zero weights for non-causal factors for each task}.

\noindent \textbf{When $p_{tc}\ne 0.5$}:

we consider the most extreme case, i.e., $p_{tc}=1$, then for Eq.\ref{eq:P_final} we have:
\begin{equation}
\begin{aligned}
    P(Y_i|{\rm A}^i,{\rm A}^j) =\frac{1}{1+e^{2Y_i\cdot \beta ^+}} 
\end{aligned}
\end{equation}
Combining Eq.\ref{eq:beta}, then we have:
\begin{equation}
\begin{aligned}
    P(Y_i|{\rm A}^i,{\rm A}^j) =\frac{1}{1+e^{2Y_i\cdot (\frac{\mu _i}{\sigma^2_i}{\rm A}^i+\frac{\mu _j}{\sigma^2_j}{\rm A}^j)}} 
\end{aligned}
\end{equation}
In this way, the optimal classifier is both for the two factors ${\rm A}^i$ and ${\rm A}^j$. That is, \emph{if $p_{tc}\ne 0.5$, the optimal classifier has non-zero weights for non-causal factors for each task}. 

In summary, Theorem 1 is certified.

\section{Pseudo Code and Pipeline}

\begin{figure*}[t]
    \centering
    \includegraphics[width=\linewidth]{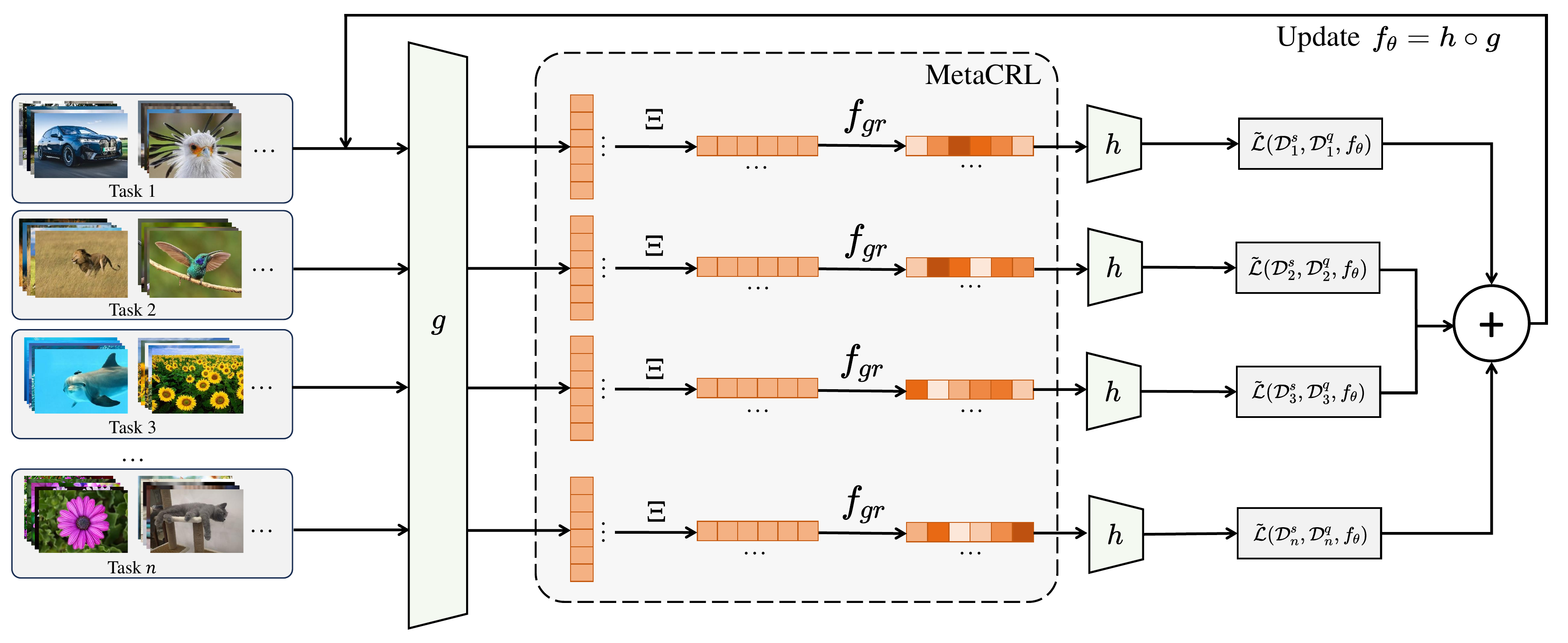}
    \caption{Overview of the meta-learning process with MetaCRL.}
    \label{fig:app_pipeline}
\end{figure*}

The pseudocode of the meta-learning process with our proposed MetaCRL is shown in Algorithm \ref{alg:algorithm}, while the pipeline of MetaCRL is shown in Figure \ref{fig:app_pipeline}. MetaCRL is a plug-and-play learner that can encode decoupled causal knowledge through the use of regularization terms. It consists of two parts: the disentanglement module and the causality module. The former aims to extract all causal generating factors and provide a subset of these factors relevant to specific tasks. The latter aims to ensure the causal relationships of the factors extracted by the disentanglement module.

\begin{algorithm*}[t]
\caption{Pseudo-Code of the meta-learning process with MetaCRL}
\label{alg:algorithm}
\textbf{Input}: Task distribution $p(\mathcal{T})$; Randomly initialize meta-learning model $f_{\theta}$ with a encoder $g$ and a classifier $h$; Randomly initialize grouping function $f_{gr}$; Initialize causal factor matrix $\Xi=\mathbb{I}^{N_z \times N_k} $\\
\textbf{Parameter}: Mini-batch $B$; Learning rates $\alpha$ and $\beta$ for the learning of $f_{\theta}$; Learning rates $\alpha_1$, $\alpha_2$, $\alpha_3$, $\alpha_4$ for the learning of $\Xi$ and $f_{gr}$; Loss weights $\lambda_1$ and $\lambda_2$\\
\textbf{Output}: meta-learning model $f_\theta$; the causal factor matrix $\Xi$ and the grouping function $f_{gr}$ of MetaCRL\\
\begin{algorithmic}[1]
\WHILE{not coverage}
\STATE Sample a batch of tasks $\mathcal{T}=\left \{ \tau_i \right \}_{i=1}^B $ from $p(\mathcal{T})$ 
\FOR{all $\tau_i$}
\STATE Sample a support set $\mathcal{D}_i^s=\left \{ (x_{i,j}^s,y_{i,j}^s) \right \}_{j=1}^{N_i^s}$ and a query set $\mathcal{D}_i^q=\left \{ (x_{i,j}^q,y_{i,j}^q) \right \}_{j=1}^{N_i^q}$
\STATE Extract causal representation for each sample using the fixed function $f_{gr}$, the fixed causal factors matrix $\Xi$, and the encoder $g$ through ${\rm{Norm[}}{f_{gr}}({\Xi ^{\rm{T}}}g({x_i}))] \odot [{\Xi ^{\rm{T}}}g(x_{i,j}^{s})]$
\STATE Update the task-specific model $f^i_{\theta}$ using the support set $\mathcal{D}_i^s$ of task $\tau_i$ with fixed $\Xi$ and $f_{gr}$ through Eq.8.
\ENDFOR
\STATE Update meta-learning model $f_{\theta}$ using all the query sets $\mathcal{D}^q$ in a single batch with fixed $\Xi$ and $f_{gr}$ through Eq.9.
\STATE Update $\Xi^{'}$ and $f_{gr}^{'}$ using all the support sets $\mathcal{D}^s$ in a single batch with fixed $f_{\theta}$ through Eq.6
\STATE Update $\Xi$ and $f_{gr}$ using all the query sets $\mathcal{D}^q$ in a single batch with fixed $f_{\theta}$ through Eq.7
\ENDWHILE
\STATE \textbf{return} solution
\end{algorithmic}
\end{algorithm*}

\section{Datasets} 

In this section, we will introduce all the datasets involved in the four scenarios of the experiment.

\subsection{Sinusoid Regression}

We choose the Sinusoid Regression problem as the first scenario in our experiment. This dataset consists of data points generated by multiple sinusoidal functions, with only a few data points for each class or pattern. Each data point includes an input value ``x", and a corresponding target output value ``y". The input values of the data points usually range within a limited scope, such as between 0 and $2\pi $.

In this experiment, we introduce noise to render the originally simple problem more challenging. Specifically, we follow the setup outlined in \cite{jiang2022role}. Each task's data is generated in the form of $A\sin w \cdot x + b + \epsilon$, where $A \in \left [ 0.1, 5.0 \right ] $, $w \in \left [ 0.5, 2.0 \right ]$, and $b \in \left [ 0,2\pi \right ] $. Next, we add Gaussian observation noise with a mean of $\mu = 0$ and a variance of $\epsilon = 0.3$ to every data point sampled from the target task. During testing, we following \cite{jiang2022role} to expand the scope of tasks by randomly sampling data. We uniformly generate $A$ values from the range $\left [ 0.1, 5.0 \right ]$, $w$ values from the range $\left [ 0.5, 2.0 \right ]$, and $b$ values from the range $\left [ 0,2\pi \right ]$. For each $(A, b)$ pair, we represent it using a one-hot vector, while w serves as the input to the network. It's worth noting that the meta-training tasks are a suitable subset of the meta-testing tasks to ensure that the meta-training process encompasses enough diversity, enabling the model to adapt well to various tasks during meta-testing.

\subsection{Image Classification}

For the second scenario, which is image classification, we select two benchmark datasets: miniImagenet \cite{vinyals2016matching,qiang2022interventional} and Omniglot \cite{lake2019omniglot,li2022metaug}. Additionally, we set up a specialized dataset named "TC," which is sampled from the motivation experiments (see ``Empirical Evidence" section for details) to evaluate the model's performance on task confounders. Next, we introduce the three datasets in this scenario.

\textbf{miniImagenet}. This dataset consists of 50,000 training images and 10,000 testing images, evenly distributed across 100 categories. Among these 100 random categories, the first 80 are used for training, and the last 20 are used for testing. It's important to note that the final 20 classes are not seen during training. All data is sampled from Imagenet.

\textbf{Omniglot}. This dataset aims to develop more human-like learning algorithms. It includes 1,623 different handwritten characters from 50 different alphabets. Each of the 1,623 characters is drawn by 20 different people using Amazon's Mechanical Turk. Each image is paired with stroke data $\left [ x, y, t \right ]$ sequences and time coordinates (t) in milliseconds.

\textbf{TC}. This dataset is sampled from tasks involving task confounders in the presence of the miniImagenet and Omniglot datasets. Specifically, we first identify cases where both test and training accuracy decrease in two sets of models from the motivation experiments, ``batchsize=10" and ``batchsize=20" ((see ``Empirical Evidence" section for details). Subsequently, we extract the $\mathcal{T}_1$ and $\mathcal{T}_2$ tasks associated with this situation, as a subset of the ``TC" dataset. A total of 50 sets of tasks are selected to compose the ``TC" dataset.

\subsection{Drug Activity Prediction}

For the third scenario, which involves drug activity prediction, we adopt the data partitioning as described in \cite{martin2019all,jiang2022role}. We sample 4276 tasks from ChEMBL \cite{gaulton2012chembl} to form the baseline dataset and preprocess it following the setup of \cite{martin2019all}.

ChEMBL is a widely used database in chemical biology and drug research, containing extensive biological activity and chemical information. It encompasses over 1.9 million compounds, more than 2 million bioactivity assay results, and thousands of biological targets. All information is stored in a structured manner, including structural information of drug compounds, results of bioactivity assays, descriptions of drug targets, and more. Following the approach in \cite{martin2019all}, we separate the training compounds in the support set and the testing compounds in the query set. The division of tasks for meta-training, meta-validation, and meta-testing is 4100, 76, and 100, respectively.

\subsection{Pose Prediction}

For the fourth scenario, we select the Pascal 3D dataset \cite{xiang2014beyond} as the benchmark dataset and process it accordingly. We randomly select 50 objects for meta-training and 15 additional objects for meta-testing.

The Pascal 3D dataset consists of outdoor images and includes 12 classes of rigid objects selected from the PASCAL VOC 2012 dataset. These objects are annotated with pose information, including azimuth, elevation, and distance to the camera. Additionally, the Pascal 3D dataset includes pose-annotated images for these 12 categories from the ImageNet dataset. We further preprocess the pose task as follows: there are 50 categories for meta-training and 15 categories for meta-testing. Each category contains 100 grayscale images with dimensions of $128 \times 128$.

\section{Baselines}
In this section, we will introduce the plug-and-play generalization baselines used in the experiments, namely MetaMix \cite{yao2021improving} and Dropout-Bins \cite{jiang2022role}, along with four meta-learning baselines, i.e., MAML \cite{finn2017model}, ANIL \cite{raghu2019rapid}, MetaSGD \cite{li2017meta}, and T-NET \cite{lee2018gradient}.

\textbf{MetaMix}. MetaMix is a baseline approach that focuses on improving generalization in meta-learning tasks. It employs techniques that enhance the model's ability to handle variations and adapt to new tasks more effectively.

\textbf{Dropout-Bins}. Dropout-Bins is another baseline strategy, involving the utilization of dropout techniques for improving generalization in the context of meta-learning. Dropout techniques are often employed to enhance model robustness and mitigate overfitting.

\textbf{MAML (Model-Agnostic Meta-Learning)}. MAML is a popular meta-learning algorithm that aims to find a model initialization that can be fine-tuned to new tasks with a small number of gradient steps. It focuses on learning a good initialization that facilitates fast adaptation.

\textbf{ANIL (Almost No Inner Loop)}. ANIL is an approach designed to reduce the number of inner loop iterations during meta-learning. It optimizes the meta-learner by reducing the reliance on costly inner loop optimization steps, aiming to achieve more efficient training.

\textbf{MetaSGD}. MetaSGD is a meta-learning algorithm that adapts the learning rate during the meta-training process. It focuses on learning to optimize the learning rate and potentially improves the model's ability to generalize across tasks.

\textbf{T-NET (Task-Agnostic Network)}. T-NET is a type of architecture that learns a shared representation across tasks. It aims to develop a task-agnostic feature extractor that captures common patterns in different tasks, contributing to improved generalization.

These methods and backbones are essential components of the experimental setup, and they are used to construct a comprehensive empirical analysis in this paper.

\begin{figure*}[t]
    \centering
    \includegraphics[width=\linewidth]{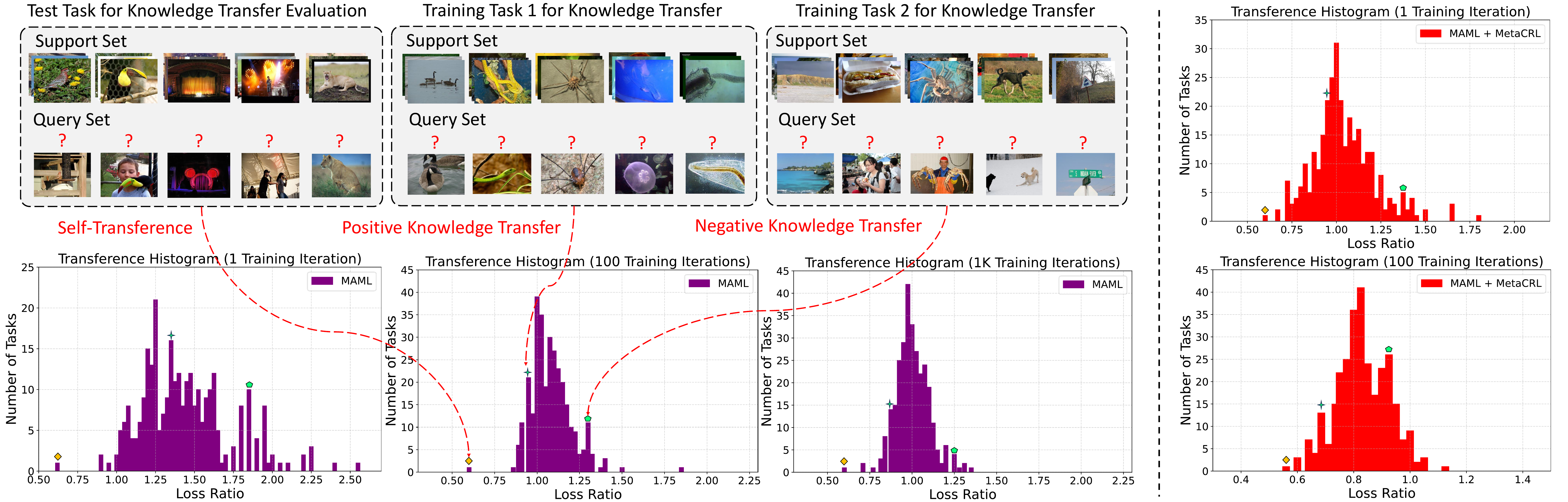}
    \caption{Knowledge transfer in miniImagenet dataset using MAML and MAML+MetaCRL as meta-learning baselines. For both positive knowledge transfer ($\mathcal{R}_{kt}<1$) and negative knowledge transfer ($\mathcal{R}_{kt}>1$), an exemplar task is shown. Here, we simply use the $\mathcal{R}_{kt}$ threshold to classify the transference of a task as positive or negative.}
    \label{app_fig:F}
\end{figure*}

\begin{figure*}[t]
    \centering
    \begin{subfigure}{0.33\textwidth}
        \centering
        \includegraphics[width=\textwidth]{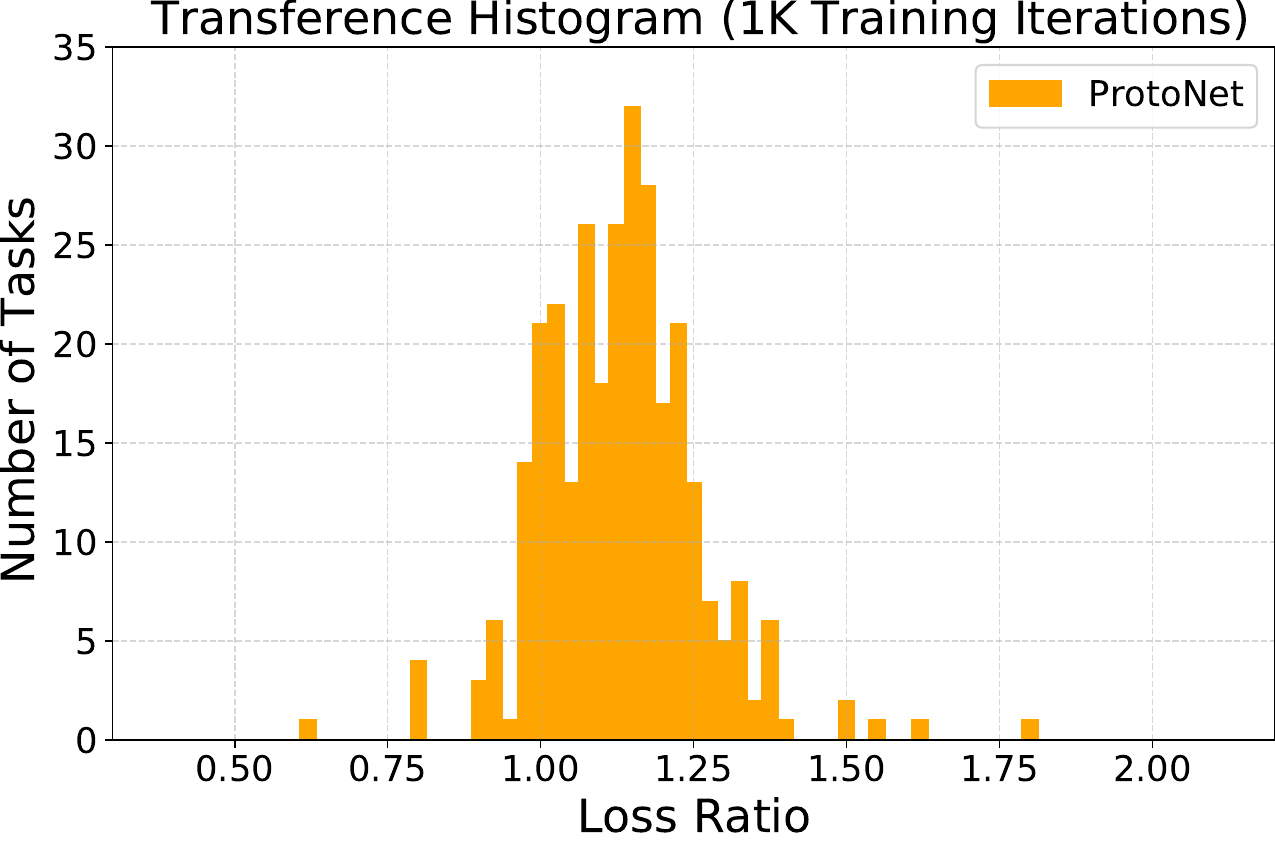}
        \caption{Knowledge transfer using ProtoNet} 
    \end{subfigure}
    \hfill
    \begin{subfigure}{0.33\textwidth}
        \centering
        \includegraphics[width=\textwidth]{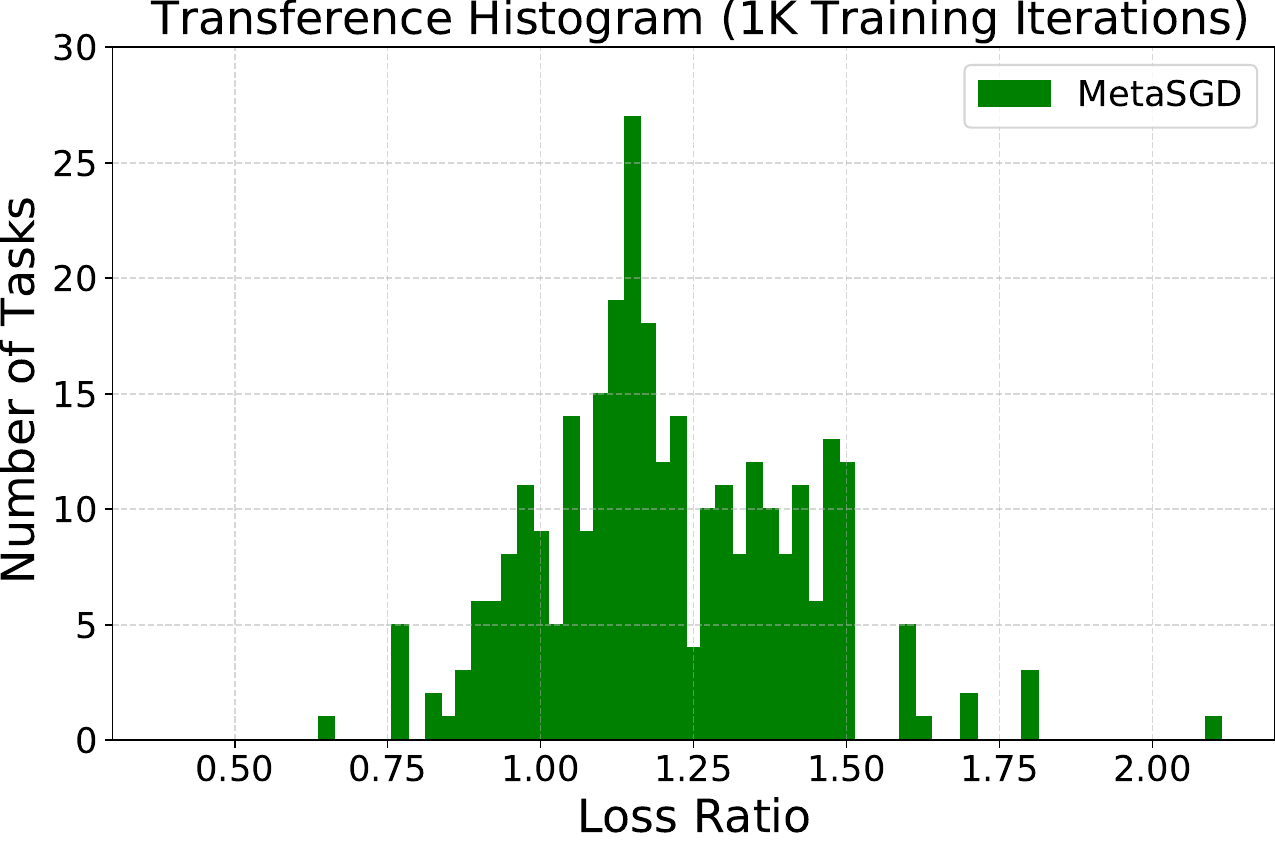}
        \caption{Knowledge transfer using MetaSGD} 
    \end{subfigure}
        \hfill
    \begin{subfigure}{0.33\textwidth}
        \centering
        \includegraphics[width=\textwidth]{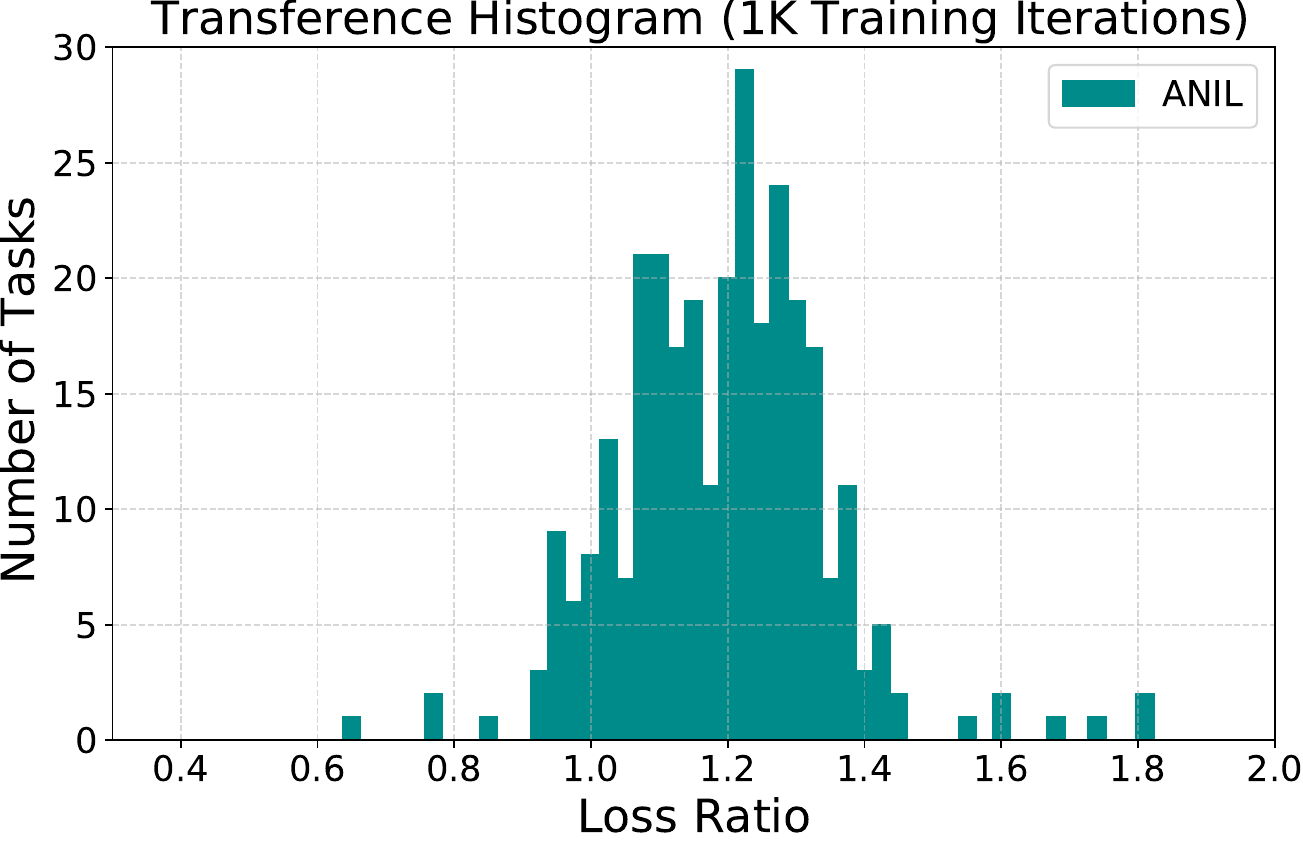}
        \caption{Knowledge transfer using ANIL} 
    \end{subfigure}
    \caption{Knowledge transfer experiments (1K Training Iterations) using ProtoNet, MetaSGD, and ANIL as meta-learning baselines using the same experimental settings in Subsection 3.2 of the main text.}
    \label{app_fig:F2}
\end{figure*}

\begin{figure*}
    \centering
    \begin{subfigure}{0.33\textwidth}
        \centering
        \includegraphics[width=\textwidth]{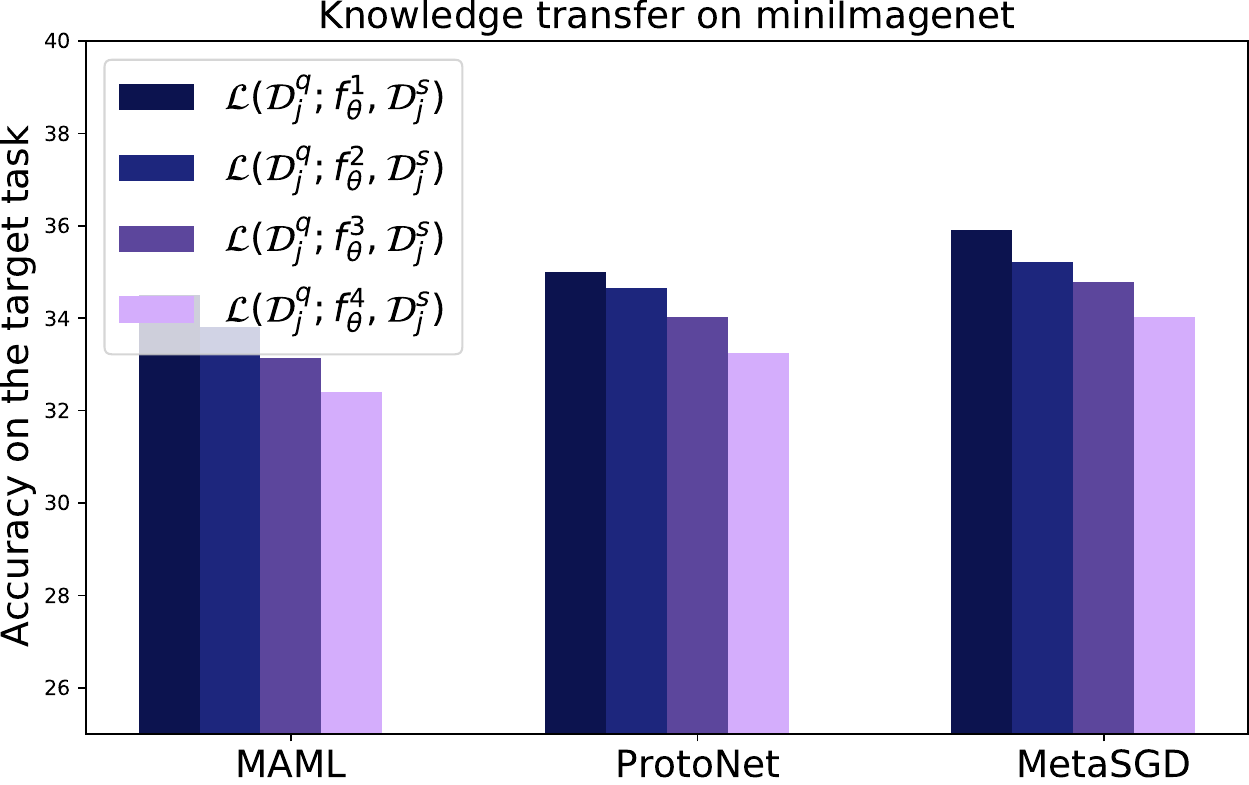}
        \caption{Knowledge transfer on miniImagenet} 
    \end{subfigure}
    \hfill
    \begin{subfigure}{0.33\textwidth}
        \centering
        \includegraphics[width=\textwidth]{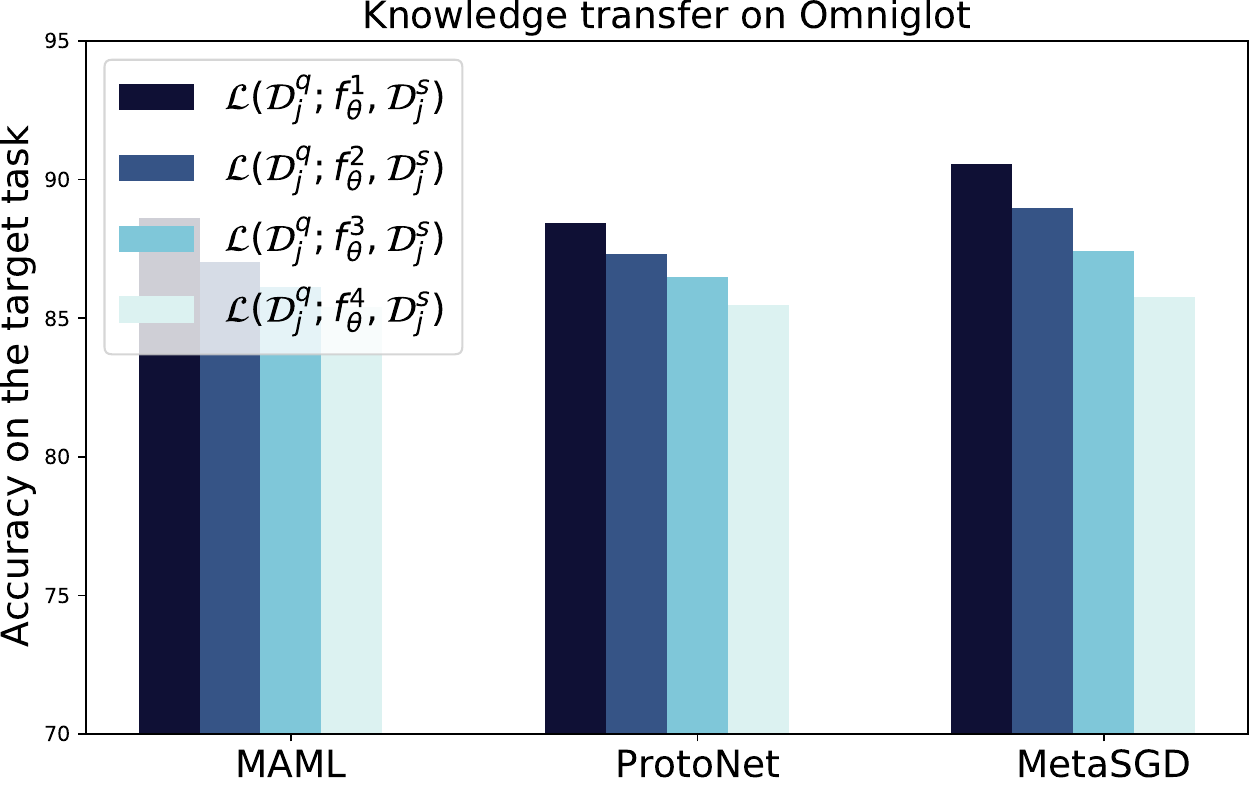}
        \caption{Knowledge transfer on Omniglot} 
    \end{subfigure}
        \hfill
    \begin{subfigure}{0.33\textwidth}
        \centering
        \includegraphics[width=\textwidth]{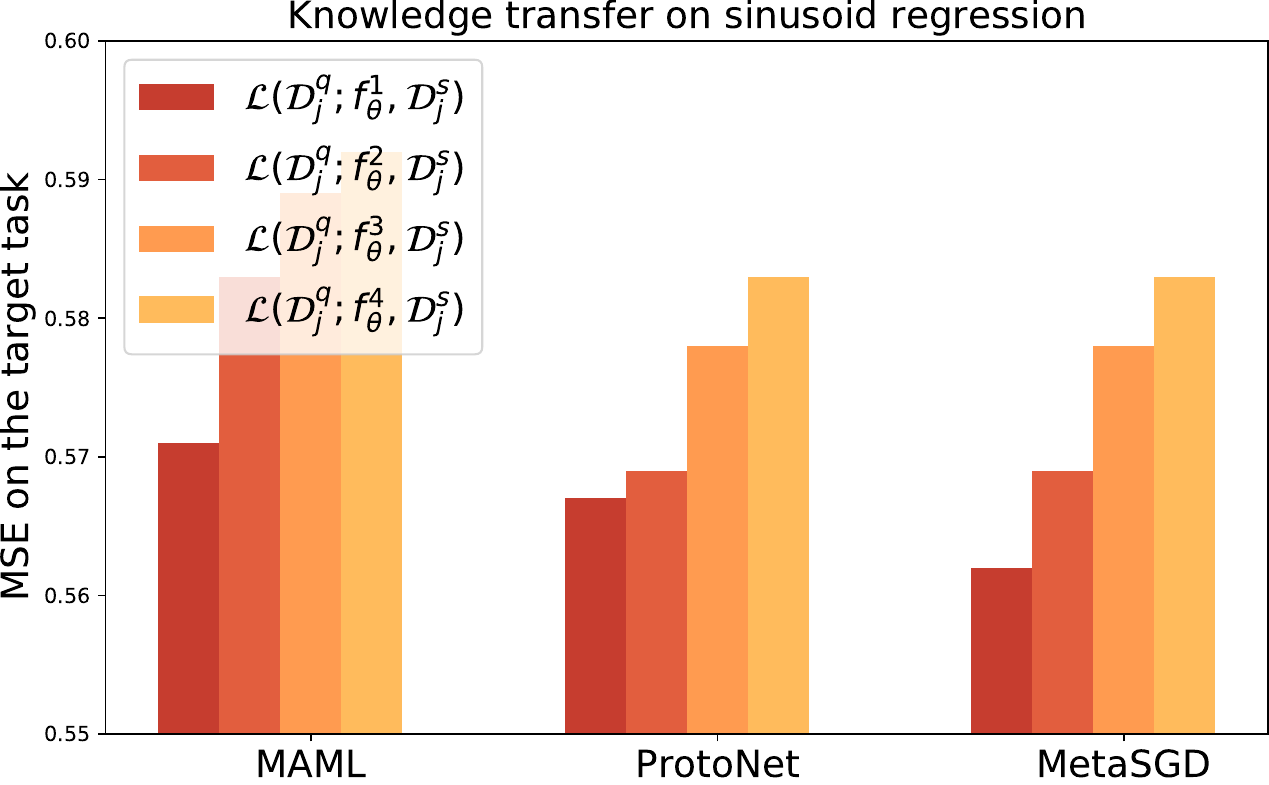}
        \caption{Knowledge transfer on sinusoid regression} 
    \end{subfigure}
    \caption{Knowledge transfer experiments for joint learning of meta-learning on miniImagenet, Omniglot, and Sinusoid Regression. These experiments evaluate the effects of multiple training tasks instead of a single training task to a single test task. By calculating the losses of different models on the target task, we can obtain the knowledge transfer effect of a set of training tasks on the target task.}
    \label{fig:app_kt}
\end{figure*}

\begin{figure*}
  \centering
  \begin{subfigure}{0.23\textwidth}
    \centering
    \includegraphics[width=\linewidth]{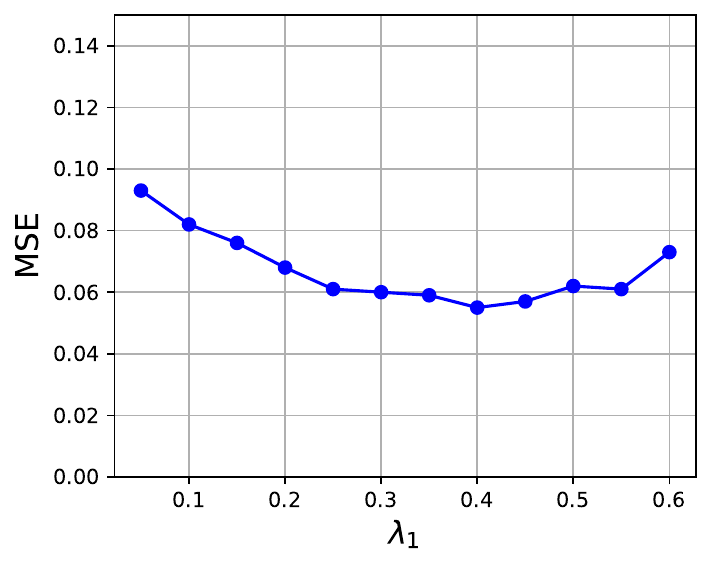}
    \caption{$\lambda_1$ on sinusoid regression}
  \end{subfigure}
  \hfill
  \begin{subfigure}{0.23\textwidth}
    \centering
    \includegraphics[width=\linewidth]{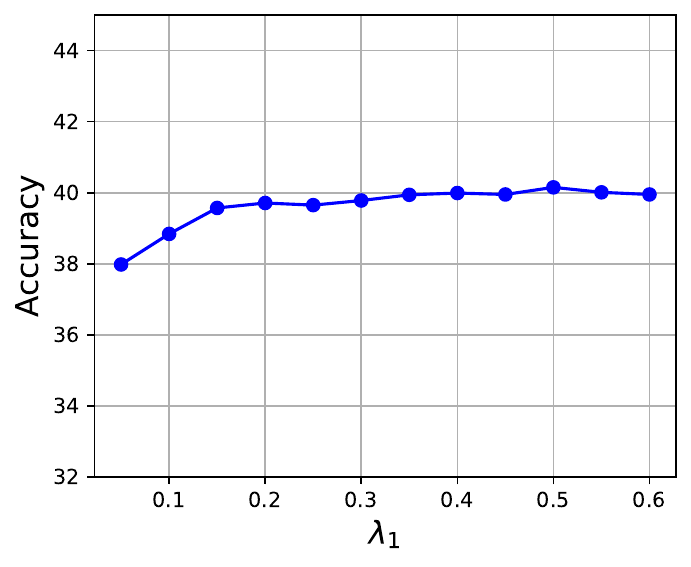}
    \caption{$\lambda_1$ on miniImagenet}
  \end{subfigure}
  \hfill
  \begin{subfigure}{0.23\textwidth}
    \centering
    \includegraphics[width=\linewidth]{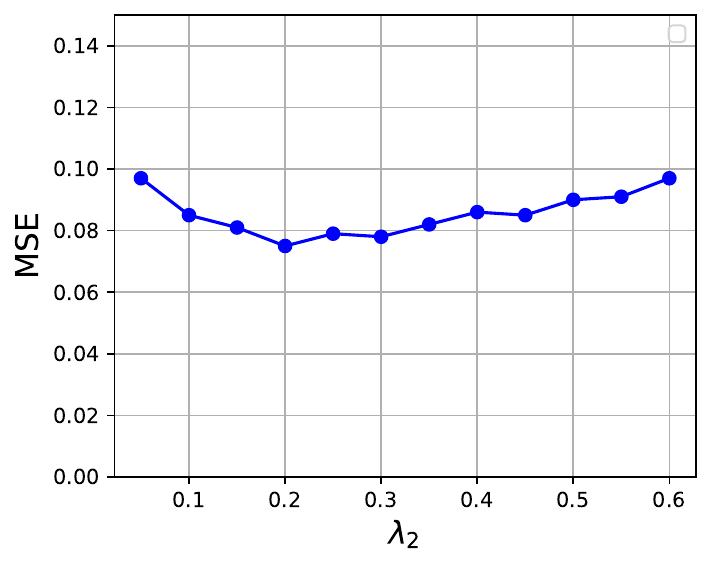}
    \caption{$\lambda_2$ on sinusoid regression}
  \end{subfigure}
  \hfill
  \begin{subfigure}{0.23\textwidth}
    \centering
    \includegraphics[width=\linewidth]{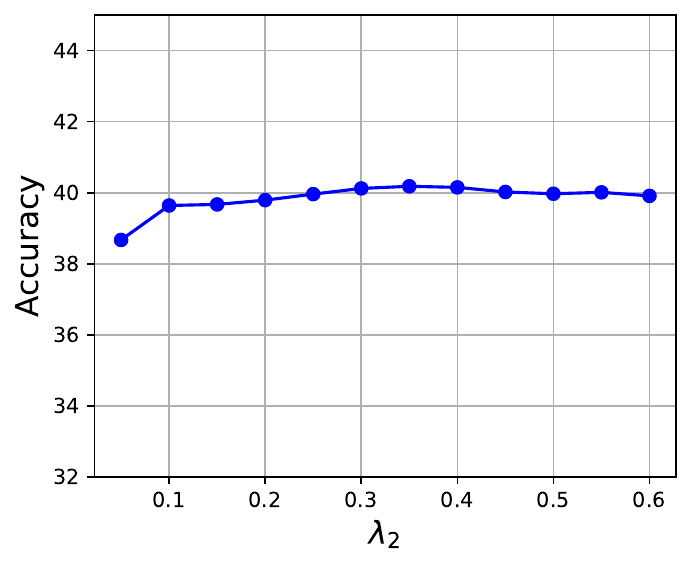}
    \caption{$\lambda_2$ on miniImagenet}
  \end{subfigure}
  \caption{Effect of hyperparameters $\lambda_1$ and $\lambda_2$ on model performance at different values. $\lambda_1$ and $\lambda_2$ are the hyperparameters of the regularization terms, i.e., $\mathcal{L}_{\rm{DM}}(\Xi) $ and $\mathcal{L}_{\rm{DM}}(f_{gr})$ of Eq.6 in the disentangling module, respectively. We conduct experiments on regression and classification scenarios, i.e., sinusoid regression and miniImagenet.}
  \label{fig:appendix1}
\end{figure*}

\section{Implementation and Architecture}

In the context of the meta-learning framework, we select the Conv4 backbone \cite{finn2017model} as the foundation for the encoders of our backbones. Following the convolution and filtering steps, we apply batch normalization, ReLU non-linear activation, and $2 \times 2$ max pooling (achieved through stride convolutions) sequentially. The final output of this encoder's last layer, which is calculated with the fixed $\Xi$ and $f_{gr}$, is then directed into a softmax layer. These network architectures undergo a pretraining phase and remain unchanged throughout the training process. It's important to note that, in line with \cite{jiang2022role}, we adopt a distinct architecture for pose prediction experiments. The fundamental model consists of a fixed encoder with three convolutional blocks, coupled with an adaptive decoder featuring four convolutional blocks. Each of these blocks encompasses a convolutional layer, a batch normalization layer, and ReLU activation.

Turning to the structure of the MetaCRL, it employs a two-layer Multilayer Perceptron (MLP) neural network, with each layer containing 500 neurons. Following every fully connected layer, there is a subsequent layer for batch normalization, alongside an activation function utilizing leaky rectified linear units (Leaky ReLU). This particular MLP serves the purpose of updating symbols $\Xi$ and $f_{gr}$, and it is encapsulated with modularity to ensure seamless integration into various meta-learning frameworks.

Moving on to the optimization process, we employ the Adam optimizer to train our model. Momentum and weight decay are set at $0.9$ and $10^{-4}$, respectively. The initial learning rate for all experiments is established at 0.4, with the flexibility for linear scaling as required. Additionally, we provide detailed explanations within the main text of each section regarding the weight of loss associated with distinct regularization terms. All experimental procedures are executed using NVIDIA RTX 4090 GPUs.

\section{Additional Results}
In this section, we first illustrate the additional experiments and the full results of Subsection 3.2 (Empirical Evidence), i.e., conduct knowledge transfer experiments under various settings, including using multiple ML baselines, using different datasets, and training on multiple tasks simultaneously. Next, we present the additional experiments and the full results of the comparison mentioned in Section 5 (Experiment), including hyperparameter sensitivity, results together with task augmentation, full results of comparison experiments, and full results of ablation studies.

\begin{table}
\begin{center}
\resizebox{\linewidth}{!}{
\begin{tabular}{l|c|c}
\toprule
\textbf{Model} & \textbf{5-shot} & \textbf{10-shot}\\
\midrule
IFSL  &  0.592 $\pm$ 0.141 & 0.178 $\pm$ 0.040 \\
Meta-Trans & 0.577 $\pm$ 0.123 & 0.140 $\pm$ 0.024 \\
Meta-Aug & 0.531 $\pm$ 0.118 & 0.103 $\pm$ 0.031 \\
MR-MAML   &  0.581 $\pm$ 0.110 & 0.104 $\pm$ 0.029 \\
\midrule
MAML  & 0.593 $\pm$ 0.120 & 0.166 $\pm$ 0.061 \\
MAML + MetaMix  & 0.476 $\pm$ 0.109 & 0.085 $\pm$ 0.024 \\
\rowcolor{orange!10}MAML + Ours  & 0.440 $\pm$ 0.079 & \textbf{0.054 $\pm$ 0.018} \\
\rowcolor{orange!10}MAML + MetaMix + Ours  & \textbf{0.441 $\pm$ 0.081} & 0.053 $\pm$ 0.019 \\
\midrule
ANIL & 0.541 $\pm$ 0.118 & 0.103 $\pm$ 0.032 \\
ANIL + MetaMix & 0.514 $\pm$ 0.106 & 0.083 $\pm$ 0.022 \\
\rowcolor{orange!10}ANIL + Ours  & \textbf{0.468 $\pm$ 0.094} & 0.081 $\pm$ 0.019 \\
\rowcolor{orange!10}ANIL + MetaMix + Ours  & \textbf{0.468 $\pm$ 0.096} & \textbf{0.083 $\pm$ 0.019} \\
\midrule
MetaSGD  & 0.577 $\pm$ 0.126 & 0.152 $\pm$ 0.044 \\
MetaSGD + MetaMix  & 0.468 $\pm$ 0.118 & 0.072 $\pm$ 0.023 \\
\rowcolor{orange!10}MetaSGD + Ours & 0.408 $\pm$ 0.071 & 0.038 $\pm$ 0.010 \\
\rowcolor{orange!10}MetaSGD + MetaMix + Ours  & \textbf{0.409 $\pm$ 0.088} & \textbf{0.041 $\pm$ 0.012} \\
\midrule
T-NET  & 0.564 $\pm$ 0.128 & 0.111 $\pm$ 0.042 \\
T-NET + MetaMix  & 0.498 $\pm$ 0.113 & 0.094 $\pm$ 0.025\\
\rowcolor{orange!10}T-NET + Ours & 0.462 $\pm$ 0.078 & 0.071 $\pm$ 0.019 \\
\rowcolor{orange!10}T-NET + MetaMix + Ours  & \textbf{0.465 $\pm$ 0.102} & \textbf{0.077 $\pm$ 0.018} \\
\bottomrule
\hline
\end{tabular}
}
\end{center}
\caption{Performance (MSE) comparison on the sinusoid regression problem. ``+ours'' means integrating MetaCRL into the existing methods, and the best results are highlighted in \textbf{bold}.}
\label{tab:E2.1}
\end{table}

\begin{table}
\begin{center}
\resizebox{\linewidth}{!}{
\begin{tabular}{l|c|c|c}
\toprule
\textbf{Model} & \textbf{Omniglot} & \textbf{miniImagenet} & \textbf{TC}\\
\midrule
MAML  & 87.15 $\pm$ 0.61 & 33.16 $\pm$ 1.70 & 0.00 \\
MAML + MetaMix  & 91.97 $\pm$ 0.51 & 38.97 $\pm$ 1.81 & +0.42 \\
\rowcolor{orange!10}MAML + Ours & 93.00 $\pm$ 0.42 & 41.55 $\pm$ 1.76 & +4.12 \\
\rowcolor{orange!10}MAML + MetaMix + Ours  & \textbf{93.04 $\pm$ 0.45} & \textbf{41.57 $\pm$ 1.74} & \textbf{+4.14} \\
\midrule
ProtoNet  & 88.51 $\pm$ 0.54 & 33.96 $\pm$ 1.64 & 0.00 \\
ProtoNet + MetaMix  & 89.97 $\pm$ 0.49 & 34.51 $\pm$ 1.51 & 0.17 \\
\rowcolor{orange!10}ProtoNet + Ours & 91.31 $\pm$ 0.60 & 35.78 $\pm$ 1.44 & 3.97 \\
\rowcolor{orange!10}ProtoNet + MetaMix + Ours  & 91.65 $\pm$ 0.56 & 36.01 $\pm$ 1.49 & 4.02 \\
\midrule
ANIL   & 89.17 $\pm$ 0.56 & 34.96 $\pm$ 1.71 & 0.00 \\
ANIL + MetaMix  & 92.88 $\pm$ 0.51 & 37.82 $\pm$ 1.75 & -0.10 \\
\rowcolor{orange!10}ANIL + Ours  & 92.91 $\pm$ 0.52 & 38.55 $\pm$ 1.81 & +3.56 \\
\rowcolor{orange!10}ANIL + MetaMix + Ours  & \textbf{92.99 $\pm$ 0.51} & \textbf{38.60 $\pm$ 1.79} & \textbf{+3.58} \\
\midrule
MetaSGD  & 87.81 $\pm$ 0.61 & 33.97 $\pm$ 0.92 & 0.00 \\
MetaSGD + MetaMix  & 93.44 $\pm$ 0.45 & 40.28 $\pm$ 0.96 & +0.05 \\
\rowcolor{orange!10}MetaSGD + Ours & \textbf{94.12 $\pm$ 0.43} & 41.22 $\pm$ 0.93 & +6.19 \\
\rowcolor{orange!10}MetaSGD + MetaMix + Ours  & 94.08 $\pm$ 0.44 & \textbf{41.24 $\pm$ 0.95} & \textbf{+6.21} \\
\midrule
T-NET  & 87.66 $\pm$ 0.59 & 33.69 $\pm$ 1.72 & 0.00 \\
T-NET + MetaMix  & 93.16 $\pm$ 0.48 & 39.18 $\pm$ 1.73 & +0.28 \\
\rowcolor{orange!10}T-NET + Ours & 93.81 $\pm$ 0.52 & 40.08 $\pm$ 1.74 & +4.65 \\
\rowcolor{orange!10}T-NET + MetaMix + Ours  & \textbf{93.91 $\pm$ 0.52} & \textbf{40.15 $\pm$ 1.74} & \textbf{+4.71} \\
\bottomrule
\hline
\end{tabular}
}
\end{center}
\caption{Performance (accuracy $\pm $ 95\% confidence interval) of image classification on (20-way 1-shot) Omniglot dataset and (5-way 1-shot) miniImagenet dataset. The ``+'' and ``-'' indicate the performance changes after adding MetaMix, MetaCRL, and MetaMix + MetaCRL. The best results are highlighted in \textbf{bold}.}
\label{tab:E2.2}
\end{table}

\begin{table*}
\begin{center}
\small
\resizebox{0.8\linewidth}{!}{
\begin{tabular}{l|cc|cc|cc}
\toprule
\multirow{2}{*}{\textbf{Model}} 
     & \multicolumn{2}{c|}{\textbf{Omniglot}} & \multicolumn{2}{c|}{\textbf{miniImagenet}} & \multirow{2}{*}{\textbf{TC}} \\ 
     & \textbf{20-way 1-shot} & \textbf{20-way 5-shot} & \textbf{5-way 1-shot} & \textbf{5-way 5-shot} & \\
\midrule
IFSL  & 88.51 $\pm$ 0.49 & 93.62 $\pm$ 0.21 & 36.21 $\pm$ 1.62 & 55.44 $\pm$ 0.95 &  $\setminus $ \\
Meta-Trans & 87.39 $\pm$ 0.51 & 92.13 $\pm$ 0.19 & 35.19 $\pm$ 1.58 & 54.31 $\pm$ 0.88 & $\setminus $ \\
Meta-Aug & 89.77 $\pm$ 0.62 & 94.56 $\pm$ 0.20 & 34.76 $\pm$ 1.52 & 54.12 $\pm$ 0.94 & $\setminus $ \\
MR-MAML & 89.28 $\pm$ 0.59 & 95.01 $\pm$ 0.23 & 35.01 $\pm$ 1.60 & 55.06 $\pm$ 0.91 & $\setminus $ \\
\midrule
MAML  &  87.15 $\pm$ 0.61 & 93.51 $\pm$ 0.25 &  33.16 $\pm$ 1.70 & 51.95 $\pm$ 0.97 & 0.00 \\
MAML + MetaMix  & 91.97 $\pm$ 0.51 & 97.95 $\pm$ 0.17 & 38.97 $\pm$ 1.81 & 58.96 $\pm$ 0.95 & +0.42 \\
MAML + Dropout-Bins & 92.89 $\pm$ 0.46 & 98.03 $\pm$ 0.15 & 39.66 $\pm$ 1.74 & 59.32 $\pm$ 0.93 & -0.14 \\
\rowcolor{orange!10}\textbf{MAML + Ours}  & \textbf{93.00 $\pm$ 0.42} &  \textbf{98.39 $\pm$ 0.15} &\textbf{41.55 $\pm$ 1.76} & \textbf{60.01 $\pm$ 0.95} & \textbf{+4.12} \\
\midrule
ProtoNet  & 89.15 $\pm$ 0.46 & 94.01 $\pm$ 0.19 & 33.76 $\pm$ 0.95 & 50.28 $\pm$ 1.31 & 0.00 \\
ProtoNet + MetaMix  & 91.08 $\pm$ 0.51 & 94.32 $\pm$ 0.29 & 34.23 $\pm$ 1.55 & 51.77 $\pm$ 0.89 & +0.28 \\
ProtoNet + Dropout-Bins & 92.13 $\pm$ 0.48 & 94.89 $\pm$ 0.23 & 34.62 $\pm$ 1.54 & 52.13 $\pm$ 0.97 & +0.03 \\
\rowcolor{orange!10}\textbf{ProtoNet + Ours}  & \textbf{93.09 $\pm$ 0.25} & \textbf{95.34 $\pm$ 0.18} & \textbf{34.97 $\pm$ 1.60} & \textbf{53.09 $\pm$ 0.93} & \textbf{+3.96} \\
\midrule
ANIL   & 89.17 $\pm$ 0.56 & 95.85 $\pm$ 0.19 & 34.96 $\pm$ 1.71 & 52.59 $\pm$ 0.96 & 0.00 \\
ANIL + MetaMix  & 92.88 $\pm$ 0.51 & 98.36 $\pm$ 0.13 & 37.82 $\pm$ 1.75 & 59.03 $\pm$ 0.93 & -0.10 \\
ANIL + Dropout-Bins  & 92.82 $\pm$ 0.49 & 98.42 $\pm$ 0.14 & 38.09 $\pm$ 1.76 & 59.17 $\pm$ 0.94 & +0.97 \\
\rowcolor{orange!10}\textbf{ANIL + Ours}  & \textbf{92.91 $\pm$ 0.52} & \textbf{98.77 $\pm$ 0.15} & \textbf{38.55 $\pm$ 1.81} & \textbf{59.68 $\pm$ 0.94} & \textbf{+3.56} \\
\midrule
MetaSGD  & 87.81 $\pm$ 0.61 & 95.52 $\pm$ 0.18 & 33.97 $\pm$ 1.34 & 52.14 $\pm$ 0.92 & 0.00 \\
MetaSGD + MetaMix  & 93.44 $\pm$ 0.45 & 98.24 $\pm$ 0.16 & 40.28 $\pm$ 1.64 & 60.19 $\pm$ 0.96 & +0.05 \\
MetaSGD + Dropout-Bins  & 93.93 $\pm$ 0.40 & 98.49 $\pm$ 0.12 & 40.31 $\pm$ 1.27 & 60.24 $\pm$ 0.91 & +1.08 \\
\rowcolor{orange!10}\textbf{MetaSGD + Ours}  & \textbf{94.12 $\pm$ 0.43} & \textbf{98.60 $\pm$ 0.15} & \textbf{41.22 $\pm$ 1.41} & \textbf{60.88 $\pm$ 0.91} & \textbf{+6.19} \\
\midrule
T-NET  & 87.66 $\pm$ 0.59 & 95.67 $\pm$ 0.20 & 33.69 $\pm$ 1.72 & 54.04 $\pm$ 0.99 & 0.00 \\
T-NET + MetaMix  & 93.16 $\pm$ 0.48 & 98.09 $\pm$ 0.15 & 39.18 $\pm$ 1.73 & 59.13 $\pm$ 0.99 & +0.28 \\
T-NET + Dropout-Bins  & 93.54 $\pm$ 0.49 & 98.27 $\pm$ 0.14 & 39.06 $\pm$ 1.72 & 59.25 $\pm$ 0.97 & +1.03 \\
\rowcolor{orange!10}\textbf{T-NET + Ours}  & \textbf{93.81 $\pm$ 0.52} & \textbf{98.56 $\pm$ 0.14} & \textbf{40.08 $\pm$ 1.74} & \textbf{59.40 $\pm$ 0.98} & \textbf{+4.65} \\
\bottomrule
\hline
\end{tabular}
}
\end{center}
\caption{Full results (accuracy $\pm 95\%$ confidence interval) of image classification on (5-way 1-shot and 5-way 5-shot) miniImagenet and (20-way 1-shot and 20-way 5-shot) Omniglot. The best results are highlighted in \textbf{bold}. The ``+'' and ``-'' indicate the performance changes after adding MetaMix, Dropout-Bins, and MetaCRL. The ``$\setminus $'' denotes that the result is not reported.}
\label{tab:F41}
\end{table*}

\begin{table*}
\begin{center}
\resizebox{\linewidth}{!}{
\begin{tabular}{l|ccc|ccc|ccc|ccc}
\toprule
\multirow{2}{*}{\textbf{Model}}
    & \multicolumn{3}{c|}{\textbf{Group 1}} & \multicolumn{3}{c|}{\textbf{Group 2}} & \multicolumn{3}{c|}{\textbf{Group 3}} & \multicolumn{3}{c}{\textbf{Group 4}} \\ 
    & \textbf{Mean} & \textbf{Med.} & \textbf{$>$ 0.3} & \textbf{Mean} & \textbf{Med.} & \textbf{$>$ 0.3} & \textbf{Mean} & \textbf{Med.} & \textbf{$>$ 0.3} & \textbf{Mean} & \textbf{Med.} & \textbf{$>$ 0.3} \\
\midrule
MAML  & 0.371 &  0.315 & 52 & 0.321 &  0.254 & 43 & 0.318 & 0.239 & 44 & 0.348 & 0.281 & 47 \\
MAML + Dropout-Bins  & 0.410 & 0.376 & 60 & 0.355 & 0.257 & 48 & 0.320 & 0.275 & 46 & 0.370 & 0.337 & 56 \\
\rowcolor{orange!10}MAML + Ours  & \textbf{0.413} & \textbf{0.378} & \textbf{61} & \textbf{0.360} & \textbf{0.261} & \textbf{50} & \textbf{0.334} & \textbf{0.282} & \textbf{51} & \textbf{0.375} & \textbf{0.341} & \textbf{59} \\
\midrule
ProtoNet  & 0.361 & 0.306 & 51 & 0.319 & 0.269 & 47 & 0.309 & 0.264 & 44 & 0.339 & 0.289 & 47 \\
ProtoNet + Dropout-Bins  & 0.391 & 0.358 & 59 & 0.336 & 0.271 & 48 & 0.314 & 0.268 & 45 & 0.376 & 0.341 & 57 \\
\rowcolor{orange!10}ProtoNet + Ours  & \textbf{0.409} & \textbf{0.398} & \textbf{62} & \textbf{0.379} & \textbf{0.292} & \textbf{52} & \textbf{0.331} & \textbf{0.300} & \textbf{52} & \textbf{0.385} & \textbf{0.356} & \textbf{59} \\
\midrule
ANIL  & 0.355 & 0.296 & 50 & 0.318 & \textbf{0.297} & \textbf{49} & 0.304 & 0.247 & 46 & 0.338 & 0.301 & 50  \\
ANIL + MetaMix  & 0.347 & 0.292 & 49 & 0.302 & 0.258 & 45 & 0.301 & 0.282 & 47 & 0.348 & 0.303 & 51  \\
ANIL + Dropout-Bins  & 0.394 & 0.321 & 53 & 0.338 & 0.271 & 48 & \textbf{0.312} & 0.284 & 46 & 0.368 & 0.297 & 50 \\
\rowcolor{orange!10}ANIL + Ours  & \textbf{0.401} & \textbf{0.339} & \textbf{57} & \textbf{0.341} & 0.277 & \textbf{49} & \textbf{0.312} & \textbf{0.291} & \textbf{48} & \textbf{0.371} & \textbf{0.305} & \textbf{53}  \\
\midrule
MetaSGD  & 0.389 & 0.305 & 50 & 0.324 & 0.239 & 46 & 0.298 & 0.235 & 41 & 0.353 & 0.317 & 52 \\
MetaSGD + MetaMix  & 0.364 & 0.296 & 49 & 0.312 & 0.267 & 48 & 0.271 & 0.230 & 45 & 0.338 & 0.319 & 51\\
MetaSGD + Dropout-Bins  & 0.390 & 0.342 & 57 & \textbf{0.358} & 0.339 & 56 & 0.316 & 0.269 & 43 & 0.360 & 0.311 & 50 \\
\rowcolor{orange!10}MetaSGD + Ours  & \textbf{0.398} & \textbf{0.295} & \textbf{59} & 0.356 & \textbf{0.340} & \textbf{59} & \textbf{0.321} & \textbf{0.271} & \textbf{44} & \textbf{0.373} & \textbf{0.324} & \textbf{55} \\
\bottomrule
\end{tabular}}
\end{center}
\caption{Full results on drug activity prediction. ``Mean", ``Mde.", and ``$> 0.3$" are the mean, the median value of $R^2$, and the number of analyzes for $R^2> 0.3$, which stands as a reliable indicator in pharmacology.}
\label{tab:F42}
\end{table*}

\begin{figure}
  \begin{minipage}{0.45\textwidth}
    \centering
    \includegraphics[width=\linewidth]{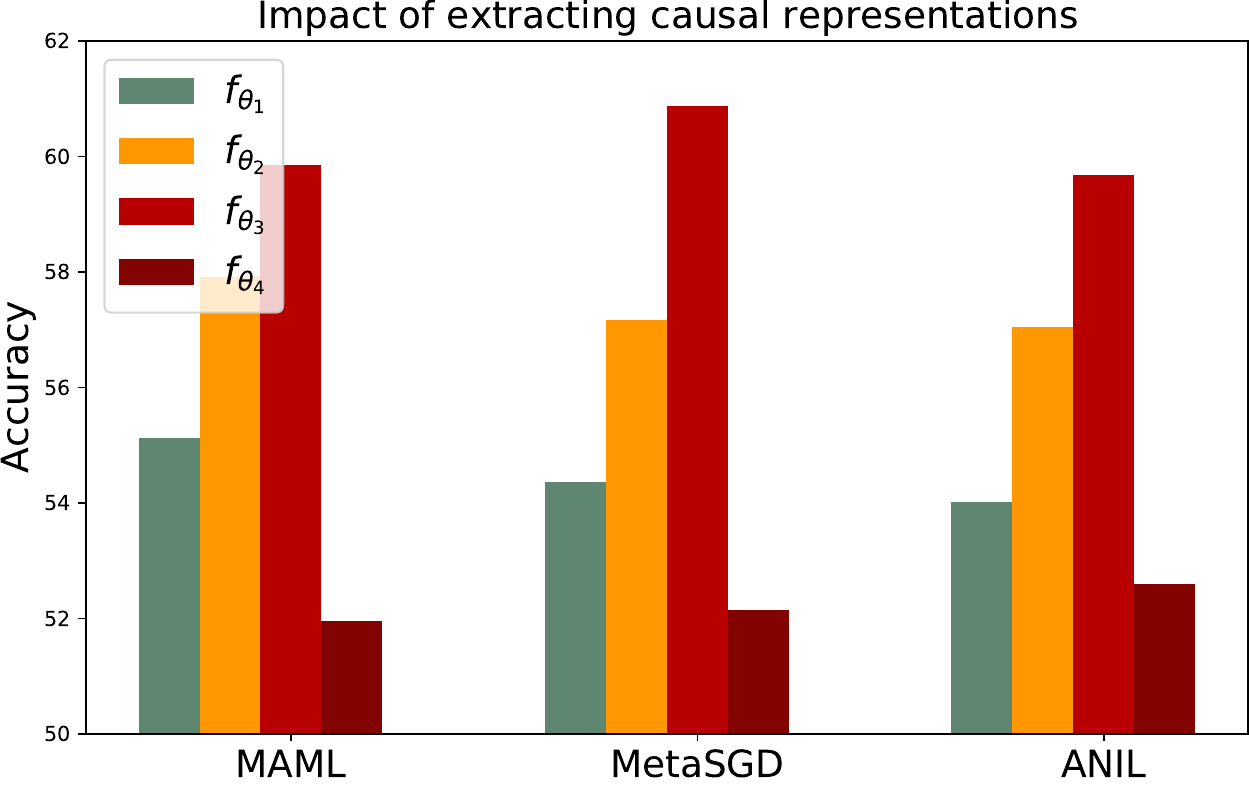}
    \caption{The accuracy of extracting causal representations. It evaluates the obtained causal representations on the results during the training process of $\Xi$ and $f_{gr}$.}
    \label{fig:app_ab_11}
  \end{minipage}
  \hfill
  \begin{minipage}{0.5\textwidth}
    \centering
    \includegraphics[width=\linewidth]{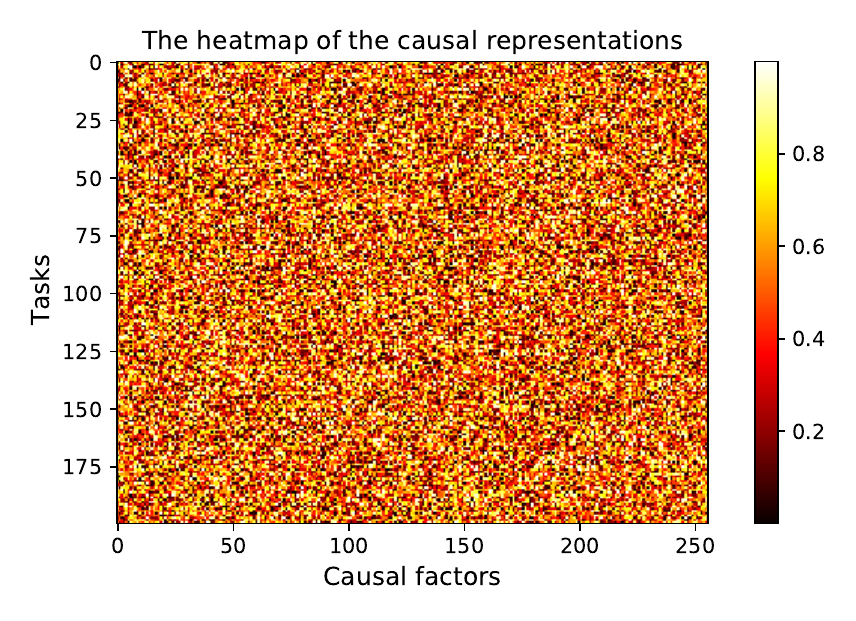}
    \vspace{-0.3in}
    \caption{The visualization of the obtained causal representations.}
    \label{fig:app_ab_12}
  \end{minipage}
  \hfill
  \begin{minipage}{0.45\textwidth}
   \vspace{0.3in}
    \includegraphics[width=\linewidth]{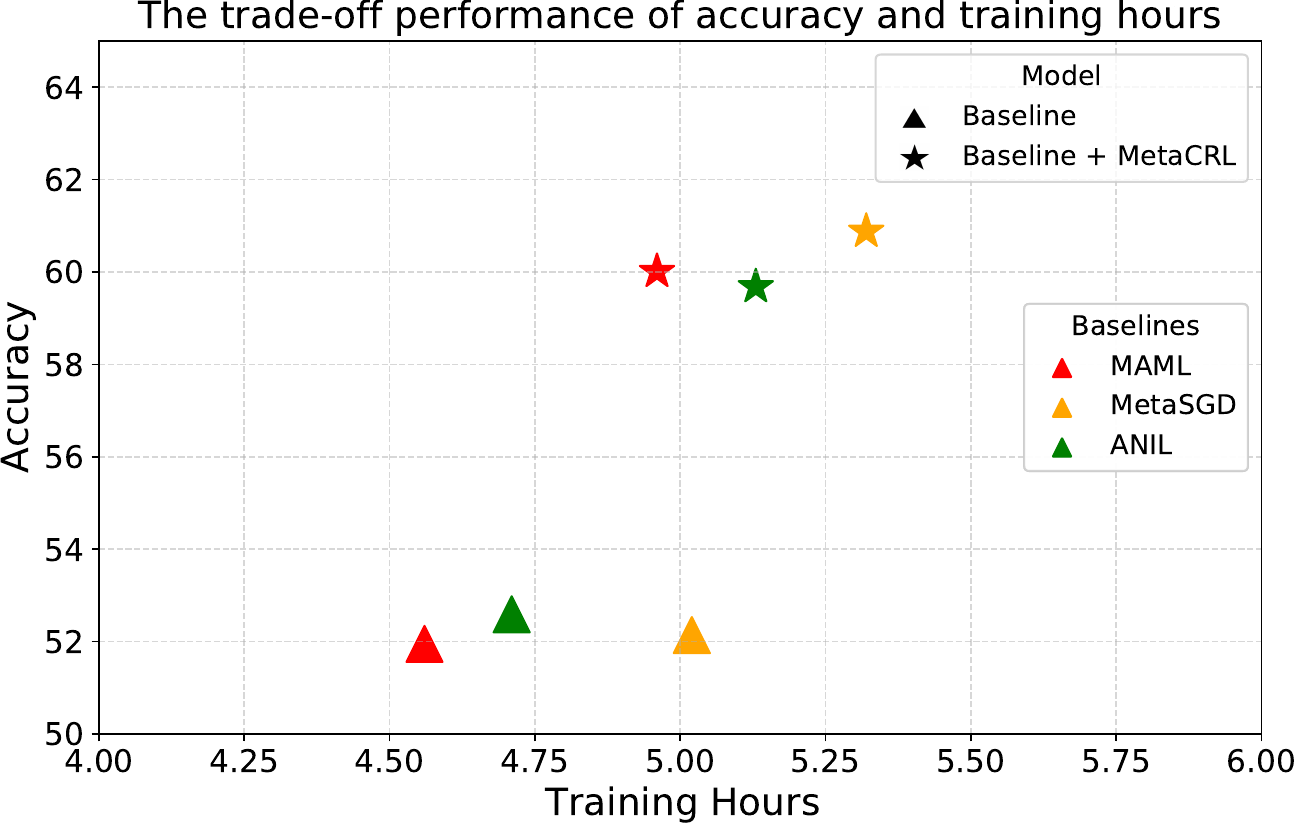}
    \caption{Model efficiency of incorporating MetaCRL, which is recorded with the same batch size and official code configuration.}
    \label{fig:app_ab_2}
  \end{minipage}
\end{figure}

\begin{table}[t]
\begin{center}
\resizebox{0.85\linewidth}{!}{
\begin{tabular}{l|c|c}
\toprule
\textbf{Model} & \textbf{10-shot} & \textbf{15-shot}\\
\midrule
IFSL  & 3.186 $\pm$ 0.256 & 2.482 $\pm$ 0.231 \\
Meta-Trans & 2.671 $\pm$ 0.248 & 2.560 $\pm$ 0.196 \\
Meta-Aug & 2.553 $\pm$ 0.265 & 2.152 $\pm$ 0.227 \\
MR-MAML  & 2.907 $\pm$ 0.255 & 2.276 $\pm$ 0.169 \\
\midrule
MAML  & 3.113 $\pm$ 0.241  & 2.496 $\pm$ 0.182 \\
MAML + MetaMix  & 2.429 $\pm$ 0.198 & 1.987 $\pm$ 0.151 \\
MAML + Dropout-Bins  & 2.396 $\pm$ 0.209 & 1.961 $\pm$ 0.134 \\
\rowcolor{orange!10}\textbf{MAML + Ours}  & \textbf{2.355 $\pm$ 0.200} & \textbf{1.931 $\pm$ 0.134} \\
\midrule
ProtoNet  & 3.571 $\pm$ 0.215 & 2.650 $\pm$ 0.210 \\
ProtoNet + MetaMix  & 3.088 $\pm$ 0.204 & 2.339 $\pm$ 0.197 \\
ProtoNet + Dropout-Bins  & 2.761 $\pm$ 0.198 & 2.011 $\pm$ 0.188 \\
\rowcolor{orange!10}\textbf{ProtoNet + Ours}  & \textbf{2.356 $\pm$ 0.171} & \textbf{1.879 $\pm$ 0.200} \\
\midrule
ANIL  & 6.921 $\pm$ 0.415 & 6.602 $\pm$ 0.385 \\
ANIL + MetaMix   & 6.394 $\pm$ 0.385 & 6.097 $\pm$ 0.311 \\
ANIL + Dropout-Bins   & 6.289 $\pm$ 0.416 & 6.064 $\pm$ 0.397 \\
\rowcolor{orange!10}\textbf{ANIL + Ours} & \textbf{6.287 $\pm$ 0.401} & \textbf{6.055 $\pm$ 0.339} \\
\midrule
MetaSGD  & 2.811 $\pm$ 0.239 & 2.017 $\pm$ 0.182 \\
MetaSGD + MetaMix  & 2.388 $\pm$ 0.204 & 1.952 $\pm$ 0.134 \\
MetaSGD + Dropout-Bins  & 2.369 $\pm$ 0.217  & 1.927 $\pm$ 0.120 \\
\rowcolor{orange!10}\textbf{MetaSGD + Ours}  & \textbf{2.362 $\pm$ 0.196} & \textbf{1.920 $\pm$ 0.191} \\
\midrule
T-NET & 2.841 $\pm$ 0.177 & 2.712 $\pm$ 0.225 \\
T-NET + MetaMix  & 2.562 $\pm$ 0.280 & 2.410 $\pm$ 0.192 \\
T-NET + Dropout-Bins  & 2.487 $\pm$ 0.212 & 2.402 $\pm$ 0.178 \\
\rowcolor{orange!10}\textbf{T-NET + Ours}  & \textbf{2.481 $\pm$ 0.274}  & \textbf{2.400 $\pm$ 0.171} \\
\bottomrule
\hline
\end{tabular}}
\end{center}
\caption{Performance (MSE $\pm $ 95\% confidence interval) comparison on pose prediction. ``+ours'' means integrating MetaCRL into the existing methods, and the best results are highlighted in \textbf{bold}.}
\label{tab:F43}
\end{table}

\subsection{Full Results of Empirical Evidence}

We construct an experiment for knowledge transfer in Subsection 3.2 of the main text, and show the visual results of knowledge transfer after introducing MetaCRL in Subsection 5.6. Specifically, to study the interactions between tasks during meta-learning training, we sample 300 meta-training tasks from miniImagenet dataset as source tasks. We then analyze the transfer performance from these tasks to unseen test tasks randomly sampled from miniImagenet using the constructed metric $\mathcal{R}_{kt}$. In this experiment, we use MAML as a meta-learner and perform analysis in the middle of training. Details of the experimental setup can be found in Appendix E. The integrated knowledge transfer results are shown in Figure \ref{app_fig:F}, which shows positive and negative knowledge transfer from the training task to the target task. It also provides example tasks for positive and negative knowledge transfer. From the results of using MAML + MetaCRL as meta-learning baseline, we can conclude that the introduction of MetaCRL can effectively eliminate the negative effect broad by negative knowledge transfer, i.e., eliminate task confounders.

In addition, we conduct knowledge transfer experiments using ProtoNet~\cite{protonet}, MetaSGD~\cite{li2017meta}, and ANIL~\cite{raghu2019rapid} as meta-learning baselines using the same experimental settings, as shown in Figure \ref{app_fig:F2}. The results demonstrate that the phenomenon of negative knowledge transfer always exists.

Considering that the training mechanism of meta-learning is to jointly learn a set of tasks in each training batch, we construct a knowledge transfer experiment under this setting to further explore the mutual influence between tasks. We use various meta-learning baselines, including MAML, ProtoNet, and MetaSGD. Specifically, for a specific test task $\tau_j$, we divide the training task into a positive knowledge transfer task set, denoted as $\mathcal{T} ^{pos}$, and a negative knowledge transfer task set, denoted as $\mathcal{T} ^{neg}$ according to the results above. Next, we refer to the most commonly used setting in miniImagenet dataset, i.e., batchsize=4, and follow the ratios of 4:0, 4:1, 3:1, and 2:2 respectively to sample four sets of training tasks from $\mathcal{T} ^{pos}$ and $\mathcal{T} ^{neg}$, denoted as $\mathcal{T}^{tr}_1-\mathcal{T}^{tr}_4$. Each task set contains multiple groups of tasks, and the positive knowledge transfer tasks in each group of tasks overlap across $\mathcal{T}^{tr}_1-\mathcal{T}^{tr}_4$. Then, we use the same experimental settings as mentioned in Subsection 3.2 (Empirical Evidence) to train ML models on these four sets of tasks respectively, and obtained $f_{\theta}^1-f_{\theta}^4$. Finally, we evaluate the performance of these four models on the target task $\tau_j$ respectively, that is, calculate the losses of $f_{\theta}^1-f_{\theta}^4$ on the query set $\mathcal{D}_j^q$ after fine-tuning once on the support set $\mathcal{D}_j^s$, denoted as $\mathcal{L} (\mathcal{D}_j^q; f_{\theta}^{\cdot},\mathcal{D}_j^s)$ where $f_{\theta}^{\cdot}$ are the above four models. We also perform the same experiment on two additional meta-learning benchmark datasets, i.e., Omniglot \cite{lake2019omniglot} also for the classification scenario and Sinusoid Regression for the regression scenario. We also adopt the same experimental settings as mentioned in Subsection 3.2 (Empirical Evidence) and the comparative experiments. The calculation results shown in Figure \ref{fig:app_kt} illustrate that the impact of negative knowledge transfer between tasks always exists and will not be offset with joint training.

\subsection{Hyperparameter Sensitivity}

We determine the hyperparameters of the regularization terms in our experiments based on the performance of multiple sets of meta-validation tasks. Specifically, for each experimental scenario, we conducted tests on the impact of different values of $\lambda_1$ and $\lambda_2$ on the performance of the MetaCRL model. The range for these values was set between $\left[0.05, 0.6\right]$. For each experiment, we recorded the optimal result selected for that scenario as the hyperparameters.

Figure \ref{fig:appendix1} shows an example that includes two scenarios, i.e., sinusoid regression and image classification. We test the model's performance with different ranges of $\lambda_1$ and $\lambda_2$ for MetaCRL on MAML. The experimental results are presented in Figure \ref{fig:appendix1}, demonstrating the method's robustness across different hyperparameter values.

\subsection{Results together with Task Augmentation}

Considering the variations across different scenarios, the effectiveness of task augmentation and the use of regularization terms differs significantly. For instance, insights from the study presented in \cite{yao2021improving} suggest that enhancing the dataset in the "Pose Prediction" scenario through augmentation can yield more effective results, possibly surpassing the sole reliance on meta-regularization techniques. As a result, we apply the proposed MetaCRL in conjunction with MetaMix to MAML, and evaluate its performance on three benchmark datasets, including Omniglot, miniImagenet, and sinusoid regression. The outcomes in Table \ref{tab:E2.1} and Table \ref{tab:E2.2} reveal that this combination exhibits greater improvements compared to the use of MetaMix alone.

\subsection{Full Results of Comparison Experiments}
We supplement the full experimental results for the image classification, drug activity prediction, and pose prediction scenarios in Table \ref{tab:F41}, Table \ref{tab:F42}, and Table \ref{tab:F43}, respectively. The experimental settings in these three scenarios are shown in the "Experiment" section of the main text and the aforementioned details of the experiment are in the appendix. The results demonstrate that, as anticipated, our method consistently attains excellent performance in all instances.

\subsection{Full Results of Ablation Studies}
In addition to the ablation experiment constructed in Section 5.5, we also construct ablation studies targeting the accuracy of extracting task-specific causal factors and model efficiency.

To explore the accuracy of extracting task-specific causal factors, we evaluate the impact of extracting causal representations on the results during the training process of $\Xi$ and $f_{gr}$. Specifically, we choose MAML+MetaCRL, MetaCRL+MetaCRL, and ANIL+MetaCRL as the baselines and choose miniImagenet dataset as the benchmark. For each baseline, we divide the training into three groups and respectively train the model based on the causal representation obtained before learning via Eq.7, after learning via Eq.7, and after learning via Eq.8, obtaining $f_{\theta_1}$, $f_{\theta_2}$, and $f_{\theta_3}$. Finally, we record the effects (5-way 5-shot accuracy) of these three models and the original meta-learning model denoted as $f_{\theta_4}$, i.e., MAML, MetaSGD, and ANIL. Among them, the training of $f_{\theta_3}$ is considered using the best causal representation, that is, the extraction of task-specific causal factors is the most accurate. If $f_{\theta_3}$ achieves the best performance, $f_{\theta_2}$ and $f_{\theta_1}$ are worse, but all higher than the original model, then the task-specific causal factors extracted by $\Xi$ and $f_{gr}$ are accurate, and their accuracy gradually increases as the training of $\Xi$ and $f_{gr}$ proceeds. The results shown in Figure \ref{fig:app_ab_11} verify this point. At the same time, we visualize the causal representation of 200 training tasks in Figure \ref{fig:app_ab_12}. From the results, we can see that the importance of the same causal factor to different tasks is different, and this importance needs to be learned. Based on these results, it shows that the task-specific causal factors extracted by MetaCRL are accurate, and the learning mechanism we designed is necessary and important.

Next, for model efficiency, we compare the trade-off performance of multiple baselines before and after using our MetaCRL on miniImagenet with Conv4 backbone. The results illustrated in Figure \ref{fig:app_ab_2} show that our MetaCRL achieves great performance with acceptable training hours.

\end{document}